%% file: posover.tex
\documentclass{article} 
\usepackage{iclr2021_conference,times}

\usepackage[hyphens]{url}
\usepackage[hidelinks]{hyperref}
\hypersetup{breaklinks=true}
\usepackage{latexsym} 

\usepackage{amsfonts}
\usepackage{bm}
\usepackage{array,amsmath}
\usepackage{amssymb}
\usepackage{dsfont}
\usepackage{float}
\usepackage{graphicx}
\usepackage{algorithm}
\usepackage{algorithmicx}
\usepackage{algpseudocode}
\usepackage{pgf,tikz}
\usepackage{tkz-euclide}
\usetikzlibrary{shapes}
\usetikzlibrary{calc,patterns,angles,quotes,backgrounds}
\usepackage{mathrsfs}
\usepackage{multirow}
\usepackage{booktabs}
\usetikzlibrary{arrows}
\usetikzlibrary{matrix}
\usepackage{threeparttable}
\usepackage[inline]{enumitem}
\usepackage[export]{adjustbox}
\usepackage{wrapfig}

\usepackage{subcaption}
\usepackage{fontawesome5}

\def\figref#1{Figure~\ref{fig:#1}}
\def\figlabel#1{\label{fig:#1}\label{p:#1}}

\def\tabref#1{Table~\ref{tab:#1}}
\def\tablabel#1{\label{tab:#1}\label{p:#1}}

\def\secref#1{\S\ref{sec:#1}}
\def\seclabel#1{\label{sec:#1}}
\def\eqref#1{Eq.~\ref{eqn:#1}}

\def\eqlabel#1{\label{eqn:#1}}

\newcommand*\diff{\mathop{}\!\mathrm{d}}

\newcounter{notecounter}
\newcommand{\enotesoff}{\long\gdef\enote##1##2{}}
\newcommand{\enoteson}{\long\gdef\enote##1##2{{
			\stepcounter{notecounter}
			{\large\bf
				\hspace{1cm}\arabic{notecounter} $<<<$ ##1: ##2
				$>>>$\hspace{1cm}}}}}
\enoteson
\enotesoff
\long\def\eat#1{}

\def\vector#1{\mathbf{#1}}
\def\matrix#1{\mathbf{#1}}
\def\tr#1{{#1}^\intercal}
\def\name#1{\emph{#1}}
\def\newcite#1{\citet{#1}}
\def\cite#1{\citep{#1}}

\def\dmax{D_{\max}}
\def\tmax#1{t_{\max}}
\newcommand{\ape}{\textsc{ape}}
\newcommand{\mam}{\textsc{mam}}
\newcommand{\pim}{position information model}
\newcommand{\Pim}{Position Information Model}
\enote{pd}{potentially add abbreviation for PIM as well}
\newcommand{\topic}{topic}
\newcommand{\Topic}{Topic}
\newcommand{\disttype}{reference point}
\newcommand{\Disttype}{Reference Point}
\newcommand{\encoding}{position encoding}
\newcommand{\Encoding}{Position Encoding}
\newcommand{\emb}{position embedding}
\newcommand{\Emb}{Position Embedding}

\title{Position Information in Transformers: \\
An Overview}
 
\author{Philipp Dufter\thanks{\mbox{\ \ } Equal contribution - random order.},  Martin Schmitt$^{*}$, Hinrich Sch\"{u}tze\\
	Center for Information and Language Processing (CIS), LMU Munich, Germany\\
	{\tt \{philipp,martin\}@cis.lmu.de}}

\iclrfinalcopy
\begin{document}

\maketitle

\input{content/content}

\bibliography{posover}
\bibliographystyle{iclr2021_conference}

\end{document}

%% file: content/content.tex
\begin{abstract}
	Transformers are arguably the main workhorse in recent Natural Language Processing research.
	By definition a Transformer is 
	invariant with respect to reordering of the input. However, language is inherently sequential and word order is essential to the semantics and syntax of an utterance. In this article, we provide an overview and theoretical comparison of existing methods to incorporate position information into Transformer models. The objectives of this survey are to \begin{enumerate*}[label=(\arabic*)]
		\item showcase that position information in Transformer is a vibrant and extensive research area;
		\item enable the reader to compare existing
                  methods by providing a unified notation
                  and
systematization of different approaches along important model dimensions;
		\item indicate what characteristics of an application should be taken into account when selecting a position encoding;
		\item provide stimuli for future research. \end{enumerate*}
\end{abstract}

\section{Introduction}
The Transformer model as introduced by \newcite{vaswani2017attention} has been found to perform well for many tasks, such as machine translation or language modeling. With the rise of pretrained language models (PLMs) \cite{peters-etal-2018-deep,howard-ruder-2018-universal,devlin-etal-2019-bert,brown2020language} Transformer models have become even more popular. As a result they are at the core of many state of the art natural language processing (NLP) models. A Transformer model consists of several layers, or blocks. Each layer is a self-attention \cite{vaswani2017attention} module followed by a feed-forward layer. Layer normalization and residual connections are additional components of a layer. 

A plain Transformer model is invariant with respect to reordering of the input. However, text data is inherently sequential. Without position information the meaning of a sentence is not well-defined, e.g., compare the sequence ``the cat chases the dog'' to the multi-set \{ the, the, dog, chases, cat \}. Clearly it should be beneficial to incorporate this essential inductive bias into any model that processes text data.

Therefore, there is a range of different methods to incorporate position information into NLP models, especially PLMs that are based on Transformer models. 
Adding position information can be done by using position
embeddings, manipulating attention matrices, or
preprocessing the input with a recurrent neural
network. Overall, there is a large variety of methods that
add 
absolute and relative position information to Transformer models. Similarly, many papers analyze and compare a subset of position embedding variants. But to the best of our knowledge, there is no broad overview of relevant work on position information in Transformers that systematically aggregates and categorizes existing approaches and analyzes the differences between them.

This survey gives an overview of existing work on incorporating and analyzing position information in Transformer models.
Concretely, we provide a theoretical comparison of over 30 Transformer position models,
and a systematization of different approaches along 
important model dimensions,
such as the number of learnable parameters,
and elucidating their differences
by means of a unified notation.
The goal of this work is not to identify the
best
way to model position information in Transformer
but rather to analyze existing works and identify common components and blind spots of current research efforts.
In summary, we aim at 
\begin{enumerate}[label=(\arabic*)]
	\item showcasing that position information in Transformer is a vibrant and extensive research area,
	\item enabling the reader to compare existing
          methods by providing a unified notation and
systematization of different approaches along important model dimensions;
	\item indicating what characteristics of an application should be taken into account when selecting a position encoding,
	\item providing stimuli for future research. 
\end{enumerate}

\section{Background}

\subsection{Notation}

Throughout this article we denote scalars with lowercase letters $x \in \mathbb{R}$, vectors with boldface $\vector{x} \in \mathbb{R}^d$, and matrices with boldface uppercase letters $\matrix{X} \in \mathbb{R}^{t \times d}$. We index vectors and matrices as follows $(x_i)_{i=1,2\dots,d} = \vector{x}$, $(X_{ij})_{i=1,2\dots,t,j=1,2,\dots d} = \matrix{X}$. Further, the $i$-th row of $\matrix{X}$ is the vector $\matrix{X}_{i} \in \mathbb{R}^{d}$. 
The transpose is denoted as $\tr{\matrix{X}}$. When we are referring to positions we use $r, s, t, \dots$ whereas we use $i, j, \dots$ to denote components of a vector. The maximum sequence length is called $\tmax{}$.

\subsection{Transformer Model}

\begin{figure}[t]
	\centering
	\begin{tikzpicture}[scale=0.6]
	\def\myheight{0.5cm}
	\def\smallheight{0.3cm}
	\node[draw=black, fill=white,rounded corners, dashed, align=center, minimum height=\smallheight, minimum width=3cm,anchor=north] at (0, 8.5)  (output) {\scriptsize \textbf{Output}};
	\node[draw=black, fill=white,rounded corners, align=center, minimum height=\smallheight, minimum width=2.5cm,anchor=north] at (0, 7.2)  (norm2) {\tiny LayerNorm};
	\node[draw=black, fill=white, rounded corners, align=center, minimum height=\smallheight, minimum width=2.5cm,anchor=north] at (0, 6.2)  (add2) {\tiny Addition};
	\node[draw=black, fill=white, rounded corners, align=center, minimum height=\myheight, minimum width=2.7cm,anchor=north] at (0, 5.3)  (ffn) {\scriptsize\textbf{Feed Forward}};
	\node[draw=black, fill=white, rounded corners, align=center, minimum height=\smallheight, minimum width=2.5cm,anchor=north] at (0, 3.7)  (norm) {\tiny LayerNorm};
	\node[draw=black, fill=white, rounded corners, align=center, minimum height=\smallheight, minimum width=2.5cm,anchor=north] at (0, 2.7)  (add) {\tiny Addition};
	\node[draw=black, fill=white, rounded corners, align=center, minimum height=\myheight, minimum width=2.7cm,anchor=north] at (0, 1.8) (attention) {\scriptsize\textbf{Attention}};
	\node[draw=black, fill=white, rounded corners, dashed, align=center, minimum height=\smallheight, minimum width=3cm,anchor=north]  at (0, 0) (input) {\scriptsize\textbf{Input}};
	\begin{scope}[on background layer]
	\draw[black,thick,rounded corners,dotted,fill=black!10] ($(output.south west)+(0.1,-0.2)$)  rectangle ($(input.north east)+(-0.1,+0.2)$);
	\draw[thick,->, to path={-- (\tikztotarget)}]	(input) to (attention);
	\draw[thick,->, to path={-- (\tikztotarget)}]	(input.west) to  [out=180,in=-90] ($(attention.west)+(-0.8,0.0)$) to [out=90,in=180] (add.west);
	\draw[thick,->, to path={-- (\tikztotarget)}]	(attention) to (add);
	\draw[thick,->, to path={-- (\tikztotarget)}]	(add) to (norm);
	\draw[thick,->, to path={-- (\tikztotarget)}]	(norm) to (ffn);
	\draw[thick,->, to path={-- (\tikztotarget)}]	(norm.west) to [bend right=-100] (add2.west);
	\draw[thick,->, to path={-- (\tikztotarget)}]	(ffn) to (add2);
	\draw[thick,->, to path={-- (\tikztotarget)}]	(add2) to (norm2);
	\draw[thick,->, to path={-- (\tikztotarget)}]	(norm2) to (output);
	\end{scope}
	\end{tikzpicture}
	\hspace{0.1cm}
	\raisebox{0.8cm}{
		\begin{tikzpicture}[scale=0.55]
		\def\myheight{0.5cm}
		\node[draw=black, fill=white, rounded corners, dashed, align=center, minimum height=\myheight, anchor=north]  at (0, -0.5) (input) {\scriptsize Input $\matrix{X}$};
		\node[draw=black, fill=white, rounded corners, align=center, minimum height=\myheight, anchor=north]  at (-2.3, 1.5) (query) {\scriptsize $\matrix{X}\matrix{W}^{(q)}$};
		\node[draw=black, fill=white, rounded corners, align=center, minimum height=\myheight, anchor=north]  at (0, 1.5) (key) {\scriptsize $\matrix{X}\matrix{W}^{(k)}$};
		\node[draw=black, fill=white, rounded corners, align=center, minimum height=\myheight, anchor=north]  at (2.3, 1.5) (value) {\scriptsize $\matrix{X}\matrix{W}^{(v)}$};
		\node[draw=black, fill=white, rounded corners, align=center, minimum height=\myheight, anchor=north]  at (-1, 3.5) (attention) {\scriptsize Attention Matrix $\matrix{A}$};
		\node[draw=black, fill=white, rounded corners, dashed, align=center, minimum height=\myheight, anchor=north]  at (0, 5.5) (output) {\scriptsize Output $\matrix{Z}$};
		\draw[thick,->, to path={-- (\tikztotarget)}]	(input.north) to  [out=90,in=-90] (query.south);
		\draw[thick,->, to path={-- (\tikztotarget)}]	(input.north) to  [out=90,in=-90] (key.south);
		\draw[thick,->, to path={-- (\tikztotarget)}]	(input.north) to  [out=90,in=-90] (value.south);
		\draw[thick,->, to path={-- (\tikztotarget)}]	(query.north) to  [out=90,in=-90] (attention.south);      
		\draw[thick,->, to path={-- (\tikztotarget)}]	(key.north) to  [out=90,in=-90] (attention.south);
		\draw[thick,->, to path={-- (\tikztotarget)}]	(attention.north) to  [out=90,in=-90] (output.south);
		\draw[thick,->, to path={-- (\tikztotarget)}]	(value.north) to  [out=90,in=-90] ($(attention.east)+(0.8,0.3)$)  to [out=90,in=-90] (output.south);
		\end{tikzpicture}}
	\caption{A rough overview of a plain Transformer
          Encoder Block (grey block) without any position
          information. The Transformer
                    Encoder Block is usually repeated
          for $l$ layers. An overview of the
          attention computation is shown on the right.}
	\figlabel{teb}
\end{figure}
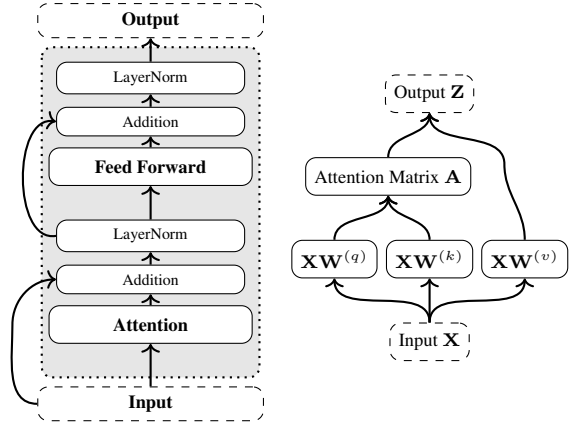

Attention mechanisms were first used in the context of machine translation by \newcite{bahdanau2015neural}. While they still relied on a recurrent neural network in its core, 
\newcite{vaswani2017attention} proposed a model that relies
on attention only. They found that it outperforms current
recurrent neural network approaches by large margins on the
machine translation task. In their paper they introduced a
new neural network architecture, the \emph{Transformer
  Model}, which is an encoder-decoder architecture. We now
briefly describe the essential building block, the \emph{Transformer Encoder Block} as shown in \figref{teb}. \figref{teb} and notation follows \cite{dufter2021distributed}. One block, also called layer, is a function $f_\theta: \mathbb{R}^{\tmax{} \times d} \to \mathbb{R}^{\tmax{} \times d}$ with $f_\theta(\matrix{X})=:\matrix{Z}$ that is defined by 
\begin{align}
\matrix{A} = \sqrt{\frac{1}{d}} \matrix{X} \matrix{W}^{(q)} \tr{(\matrix{X} \matrix{W}^{(k)})}\nonumber\\
\matrix{M} = \text{SoftMax} (\matrix{A}) \matrix{X} \matrix{W}^{(v)} \nonumber\\
\matrix{O} = \text{LayerNorm}_1(\matrix{M} + \matrix{X})\\
\matrix{F} = \text{ReLU}(\matrix{O} \matrix{W}^{(f_1)} + \vector{b}^{(f_1)}) \matrix{W}^{(f_2)} + \vector{b}^{(f_2)} \nonumber\\
\matrix{Z} = \text{LayerNorm}_2(\matrix{O} + \matrix{F})\nonumber
\end{align}
Here, $\text{SoftMax}(\matrix{A})_{ts} = e^{\matrix{A}_{ts}} / \sum_{k=1}^{\tmax{}} e^{\matrix{A}_{tk}}$ is the row-wise softmax function, $\text{LayerNorm}(\matrix{X})_{t} = \vector{g} \odot (\matrix{X}_{t} - \mu(\matrix{X}_{t}) / \sigma(\matrix{X}_{t}) + \vector{b}$ is layer normalization \cite{lei2016layer} where $\mu(\vector{x}), \sigma(\vector{x})$ returns the mean, standard deviation of a vector, 
and  $\text{ReLU}(\matrix{X}) = \max(0, \matrix{X})$ is the maximum operator applied componentwise. Note that for addition of a vector to a matrix we assume broadcasting as implemented in NumPy \cite{harris2020array}.
Overall the parameters of a single layer are
\begin{align}
\theta=(\matrix{W}^{(q)},\matrix{W}^{(k)},\matrix{W}^{(v)} \in \mathbb{R}^{d \times d}, 
\vector{g}^{(1)}, \vector{g}^{(2)}, \vector{b}^{(1)}, \vector{b}^{(2)} \in \mathbb{R}^{d}, \\
\matrix{W}^{(f_1)} \in \mathbb{R}^{d \times d_f}, \matrix{W}^{(f_2)} \in \mathbb{R}^{d_f \times d}, 
\vector{b}^{(f_1)} \in \mathbb{R}^{d_f}, \vector{b}^{(f_2)} \in \mathbb{R}^{d}), \nonumber
\end{align}
with $d$ the hidden dimension, $d_f$ the intermediate dimension, and $\tmax{}$ the maximum sequence length. 
It is common to consider multiple, say $h$, attention heads. More specifically,  $\matrix{W}^{(q)},\matrix{W}^{(k)},\matrix{W}^{(v)} \in \mathbb{R}^{d \times d_h}$ where $d = h d_h$. Subsequently, the matrices $\matrix{M}^{(h)} \in \mathbb{R}^{\tmax{} \times d_h}$ from each attention head are concatenated along their second dimension to obtain $\matrix{M}$.
A full \emph{Transformer model} is then the function $T: \mathbb{R}^{\tmax{} \times d} \to \mathbb{R}^{\tmax{} \times d} $ that consists of the composition of multiple, say $l$ layers, i.e.,  $T(\matrix{X}) = f_{\theta^l} \circ f_{\theta^{l - 1}} \circ \dots \circ  f_{\theta^{1}}(\matrix{X})$.

When considering an input $U=(u_1, u_2, \dots, u_t)$ that
consists of $t$ unit (such as characters, subwords, words)  embeddings $\matrix{U} \in \mathbb{R}^{\tmax{} \times d}$ are created by a lookup in the embedding matrix $\matrix{E} \in \mathbb{R}^{n \times d}$ with $n$ being the vocabulary size. More specifically, $\matrix{U}_i=\matrix{E}_{u_i}$ is the embedding vector that corresponds to the unit $u_i$. Finally, the matrix $\matrix{U}$ is then (among others) used as input to the Transformer model. In the case that $U$ is shorter or longer than $\tmax{}$, it is padded, i.e., filled with special \textsc{pad} symbols, or truncated.

\subsection{Order Invariance}

If we take a close look at the Transformer model, we see that 
it is invariant to reordering of the input.
More specifically, consider any permutation matrix $\matrix{P}_\pi \in \mathbb{R}^{\tmax{} \times \tmax{}}$.
When passing $\matrix{P}_\pi\matrix{X}$ to a Transformer
layer, one gets $\matrix{P}_\pi \text{SoftMax} (\matrix{A})
\tr{\matrix{P}_\pi}  \matrix{P}_\pi \matrix{X}
\matrix{W}^{(v)} = \matrix{P}_\pi \matrix{M}$, as $
\tr{\matrix{P}_\pi}  \matrix{P}_\pi$ is the identity
matrix.
All remaining operations are position-wise and thus $\matrix{P}_\pi T (\matrix{X}) = T(\matrix{P}_\pi\matrix{X})$ for any input $\matrix{X}$. As language is inherently sequential it is desirable to have $\matrix{P}_\pi T (\matrix{X}) \not = T(\matrix{P}_\pi\matrix{X})$, which can be achieved by incorporating position information.

\subsection{Encoder-Decoder}

There are different setups for using a Transformer model. One common possibility is to have an encoder only. For example, BERT \cite{devlin-etal-2019-bert} uses a Transformer model $T(\matrix{X})$ as encoder to perform masked language modeling. In contrast, a traditional sequence-to-sequence approach can be materialized by adding a decoder.
The decoder works almost identically to the encoder with two exceptions:
\begin{enumerate*}[label=(\arabic{*})]
	\item The upper triangle of the attention matrix $\matrix{A}$ is usually masked in order to block information flow from future positions during the decoding process.
	\item The output of the encoder is integrated through a cross-attention layer inserted before the feed forward layer.
\end{enumerate*}
See \cite{vaswani2017attention} for more details.
The differences between an encoder and encoder-decoder architecture are mostly irrelevant for the injection of position information and many architectures rely just on encoder layers. Thus for the sake of simplicity we will talk about Transformer Encoder Blocks in general for the rest of the article. See \secref{decoder} for position encodings that are tailored for encoder-decoder architectures.

\section{Recurring Concepts in \Pim s}

While there is a variety of approaches to integrate position information into Transformers, there are some recurring ideas, which we outline here.

\subsection{Absolute vs.\ Relative \Encoding{}}

\emph{Absolute positions} encode the absolute position of a
unit within a sentence. Another approach is to encode the
position of a unit \emph{relative} to other units. This
makes intuitively sense, as in sentences like ``The cat
chased the dog.'' and ``Suddenly, the cat chased the dog.''
the change in absolute positions due to the added word
``Suddenly'' causes only a small semantic change whereas the relative positions of ``cat'' and ``dog'' are decisive for the meaning of the sentences. 

\subsection{Representation of Position Information}

\textbf{Adding Position Embeddings (\ape).} One common approach is to add position embeddings to the input before it is fed to the actual Transformer model: If $\matrix{U} \in \mathbb{R}^{\tmax{} \times d}$ is the matrix of unit embeddings, a matrix $\matrix{P} \in \mathbb{R}^{\tmax{} \times d}$ representing the position information is added, i.e., their sum is fed to the Transformer model: $T(\matrix{U} + \matrix{P})$. For the first Transformer layer, this has the following effect:
\begin{align}
\matrix{\tilde{A}} = \sqrt{\frac{1}{d}} (\matrix{U} + \matrix{P}) \matrix{W}^{(q)} \tr{\matrix{W}^{(k)}}\tr{(\matrix{U} + \matrix{P})} \nonumber\\
\matrix{\tilde{M}} = \text{SoftMax} (\matrix{\tilde{A}}) (\matrix{U} + \matrix{P})\matrix{W}^{(v)}\nonumber\\
\matrix{\tilde{O}} = \text{LayerNorm}_2(\matrix{\tilde{M}} + \matrix{U} + \matrix{P})\\
\matrix{\tilde{F}} = \text{ReLU}(\matrix{\tilde{O}} \matrix{W}^{(f_1)} + \vector{b}^{(f_1)}) \matrix{W}^{(f_2)} + \vector{b}^{(f_2)} \nonumber\\
\matrix{\tilde{Z}} = \text{LayerNorm}_1(\matrix{\tilde{O}} + \matrix{\tilde{F}}) \nonumber
\end{align}

\begin{figure}
	\centering
\begin{tikzpicture}
\matrix (text) [matrix of nodes]{
		You \\
	are \\
	great  \\
};
\matrix (att) [matrix of math nodes,
column sep =1mm,
row sep =1mm,
left delimiter={[},
right delimiter={]},
right=0.1cm of text,
anchor=west]
{
	a_{11} & a_{12}  & a_{13}  \\
	a_{21} & a_{22}  & a_{23}  \\
	a_{31} & a_{32}  & a_{33}  \\
};
\matrix (texttop) [matrix of nodes,
above of=att]{
	You &
	are &
	great \\
};
\node[below=0.2cm of att] {\textit{Attention Matrix}};
\matrix (abs) [matrix of math nodes,
column sep =1mm,
row sep =1mm,
left delimiter={[},
right delimiter={]},
right=1cm of att,
anchor=west]
{
	p_{11} & p_{12}  & p_{13}  \\
	p_{21} & p_{22}  & p_{23}  \\
	p_{31} & p_{32}  & p_{33}  \\
};
\node[below=0.2cm of abs] {\textit{Absolute Position Bias}};
\matrix (rel) [matrix of math nodes,
column sep =1mm,
row sep =1mm,
left delimiter={[},
right delimiter={]},
right=1cm of abs,
anchor=west]
{
	r_{0} & r_{1}  & r_{2}  \\
	r_{-1} & r_{0}  & r_{1}  \\
	r_{-2} & r_{-1}  & r_{0}  \\
};
\node[below=0.2cm of rel] {\textit{Relative Position Bias}};
\end{tikzpicture}
	\caption{Example of absolute and relative position biases that can be added to the attention matrix. \textit{Left:} attention matrix for an example sentence. \textit{Middle:} learnable absolute position biases. \textit{Right:} position biases with a relative reference point. They are different from absolute encodings as they exhibit an intuitive weight sharing pattern.}
\figlabel{biases}
\end{figure}

\textbf{Modifying Attention Matrix (\mam{})}. Instead of adding position embeddings, other approaches directly modify the attention matrix. For example, by adding absolute or relative position biases to the matrix, see \figref{biases}. In fact, one big effect of adding position embeddings is that it modifies the attention matrix as follows

\begin{align}\eqlabel{posfact}
\matrix{\hat{A}} \sim
\underbrace{\matrix{U}  \matrix{W}^{(q)}\tr{\matrix{W}^{(k)}} \tr{\matrix{U}}}_{\text{unit-unit }\sim \matrix{A}}  + 
\underbrace{\matrix{P} \matrix{W}^{(q)} \tr{\matrix{W}^{(k)}}  \tr{\matrix{U}} 
	+  \matrix{U}\matrix{W}^{(q)} \tr{\matrix{W}^{(k)}}  \tr{\matrix{P}}}_{\text{unit-position}} + 
\underbrace{\matrix{P} \matrix{W}^{(q)} \tr{\matrix{W}^{(k)}} \tr{\matrix{P}}}_{\text{position-position}}
\end{align}
As indicated, the matrix $\matrix{A}$ can then be decomposed into unit-unit interactions as well as unit-position and position-position interactions. We write $\sim$ as we omit the scaling factor for the attention matrix for simplicity. 

As adding position embeddings (i.e., APE) results in a modification of
the attention matrix, APE and MAM are highly interlinked. Still, we make a distinction between these two approaches for two reasons:
\begin{enumerate*}[label=(\arabic{*})]
	\item While adding position embeddings results, among other effects, in a modified attention matrix, \mam{} \emph{only} modifies the attention matrix. 
	\item \ape{} involves learning embeddings for position information whereas \mam{} is often interpreted as adding or multiplying scalar biases to the attention matrix $\matrix{A}$, see \figref{biases}.
\end{enumerate*}

\subsection{Integration}
In theory, there are many possibilities for injecting position information, but in practice the information is either integrated in the input, at each attention matrix, or directly before the output. When adding position information at the beginning, it only affects the first layer and has to be propagated to upper layers indirectly. Often,  \ape{} is only added
at the beginning, and \mam{} approaches are used for each layer and attention head.

\section{Current \Pim s}

In this section we provide an overview of current \pim s. Note that we use the term \emph{\pim{}} to refer to a method that integrates position information, the term \emph{\encoding{}} refers to a position ID associated to units, e.g., numbered from $0$ to $t$, or assigning relative distances. A \emph{\emb{}} then refers to a numerical vector associated with a position encoding. We systematize \pim s along two dimensions: \emph{\disttype{}} and \emph{\topic{}}, see \tabref{overview}. The following sections deal with each \topic{} and within each topic we discuss approaches with different \disttype{}s. \tabref{comparison} provides more details for each method and aims at making comparisons easier.

\begin{table}[t]
	\footnotesize
	\centering
	\def\mysep{0.08cm}
	\def\topicw{1.7cm}
	\begin{tabular}{@{\hspace{\mysep}}l@{\hspace{\mysep}}
						@{\hspace{\mysep}}l@{\hspace{\mysep}}
						@{\hspace{\mysep}}l@{\hspace{\mysep}}
						@{\hspace{\mysep}}l@{\hspace{\mysep}}
						@{\hspace{\mysep}}l@{\hspace{\mysep}}						
		}
		\toprule
		&&\multicolumn{3}{c}{\textbf{\Disttype{}}}\\
		&&	\textbf{Absolute}	&	\textbf{Absolute \& Relative }	&	\textbf{Relative}	\\
		\cmidrule{2-5}
		\multirow{24}{0.3cm}{\rotatebox{90}{\textbf{\Topic{}}}}
&	\multirow{7}{\topicw}{\hyperref[sec:seq]{\textbf{Sequential}}}        	&	\cite{devlin-etal-2019-bert}	&	\cite{shaw2018self}	&	\cite{dai-etal-2019-transformer}	\\
&		&	\cite{kitaev2020reformer}	&	\cite{ke2020rethinking}	&	\cite{raffel2020exploring}	\\
&		&	\cite{liu2020learning}	&	\cite{dufter2020increasing}	&	\cite{wu2020datransformer}	\\
&		&	\cite{press2020shortformer}	&	\cite{he2020deberta}	&	\cite{huang2020improve}	\\
&		&	\cite{wang2020encoding}	&		&	\cite{shen2018disan}	\\
&		&	\cite{dehghani2019universal}	&		&	\cite{neishi2019relation}	\\
&		&		&		&	\cite{chang-etal-2021-convolutions} \\
&&&&\cite{liutkus-etal-2021-relative}	\\\cmidrule{2-5}
&	\multirow{3}{\topicw}{\hyperref[sec:sinus]{\textbf{Sinusoidal}}}        	&	\cite{vaswani2017attention}	&		&	\cite{yan2019tener}	\\
&		&	\cite{li2019augmented}	&		&	\cite{su2021roformer}	\\
&		&	\cite{likhomanenko2021cape}	&		&		\\\cmidrule{2-5}
&	\multirow{3}{\topicw}{\hyperref[sec:graphs]{\textbf{Graphs}}}        	&	\cite{shiv2019novel}	&	\cite{wang2019self}	&	\cite{zhu-etal-2019-modeling}	\\
&		& \cite{dwivedi2021generalization}	& \cite{zhang-etal-2020-graph} &	\cite{cai-lam-2020-graph}	\\
&		&		&		&	\cite{schmitt2021modeling}	\\\cmidrule{2-5}
&	\multirow{3}{\topicw}{\hyperref[sec:decoder]{\textbf{Decoder}}}        	&	\cite{takase-okazaki-2019-positional}	&		&		\\
&		&	\cite{oka-etal-2020-incorporating}	&		&		\\
&		&	\cite{bao2019non}	&		&		\\\cmidrule{2-5}
&	\multirow{4}{\topicw}{\hyperref[sec:crosslingual]{\textbf{Crosslingual}}}        	&	\cite{artetxe2020crosslingual}	&		&		\\
&		&	\cite{ding2020self}	&		&		\\
&		&	\cite{liu2020improving}	&		&		\\
&		&	\cite{liu2020do}	&		&		\\\cmidrule{2-5}
&	\multirow{3}{\topicw}{\hyperref[sec:analysis]{\textbf{Analysis}}}        	&	\cite{yang2019assessing}	&	\cite{rosendahl2019analysis}	&		\\
&		&	\cite{wang2020what}	&	\newcite{wang2021on}	&		\\
&		&		&	\cite{chen2021demystifying}	&		\\	\bottomrule																	
	\end{tabular}
	\caption{Overview and categorization of papers dealing with position information. We categorize along two dimensions: \topic{}, i.e., a tag that describes the main topic of a paper, and which \disttype{} is used for the position encodings.}
	\tablabel{overview}
\end{table}

\subsection{Sequential}
\seclabel{seq}

\subsubsection{Absolute \Encoding s}
\seclabel{seq:abs}

The original Transformer paper
 considered absolute \encoding s.
One of the two approaches proposed by \newcite{vaswani2017attention}
follows \newcite{gehring2017convolutional}
and learns a position embedding matrix $\matrix{P} \in \mathbb{R}^{t_{\text{max}} \times d}$
corresponding to the absolute positions $1, 2, \dots, \tmax{}-1, \tmax{}$ in a sequence.
This matrix is simply added to the unit embeddings $\matrix{U}$ before they are fed to the Transformer model (\ape).

In the simplest case, the \emb s are randomly initialized and then adapted during training of the network \cite{gehring2017convolutional,vaswani2017attention,devlin-etal-2019-bert}.
\newcite{gehring2017convolutional} find that adding \emb s only help marginally in a convolutional neural network. A Transformer model without any position information, however, performs much worse for some tasks (e.g., \citealp{wang2019self}, \citealp{wang2021on}).

\begin{figure}[t]
	\centering
	\begin{tikzpicture}[scale=0.6]
	\scriptsize
	\def\aheight{1cm};
	\def\bheight{1cm};
	\def\mywidth{0.5cm};
	\node[minimum height=\aheight, minimum width=\mywidth, anchor=west] at (0, 0) (x0) {};
	\foreach \i in {1,...,3}
	{
		\pgfmathtruncatemacro{\current}{(\i)};
		\pgfmathtruncatemacro{\previous}{(\i - 1)}
		\node[minimum height=\aheight, minimum width=\mywidth, anchor=west] at (x\previous.east) (x\current) {$\matrix{P}^{(1)}_{\current}$};
		\node[minimum height=\aheight, minimum width=\mywidth, anchor=north] at (x\current.south) (y\current) {$\matrix{P}^{(2)}_{1}$};
		\draw[black,dashed] ($(x\current.north east)+(-0.0,0.0)$) -- ($(y\current.south east)+(+0.0,-0.0)$);
	}
	\node[minimum height=\aheight, minimum width=\mywidth, anchor=east] at (x3.east) (z0) {};
	\foreach \i in {1,...,3}
	{
		\pgfmathtruncatemacro{\current}{(\i)};
		\pgfmathtruncatemacro{\previous}{(\i - 1)}
		\node[minimum height=\aheight, minimum width=\mywidth, anchor=west] at (z\previous.east) (z\current) {$\matrix{P}^{(1)}_{\current}$};
		\node[minimum height=\aheight, minimum width=\mywidth, anchor=north] at (z\current.south) (w\current) {$\matrix{P}^{(2)}_{2}$};
		\draw[black,dashed] ($(z\current.north east)+(-0.0,0.0)$) -- ($(w\current.south east)+(+0.0,-0.0)$);
	}
	\node[minimum height=\aheight, minimum width=\mywidth, anchor=west] at (z3.east) (zdots) {...};
	\node[minimum height=\aheight, minimum width=\mywidth, anchor=west] at (w3.east) (wdots) {...};
	\draw[black] ($(x1.north west)+(-0.1,0.1)$)  rectangle ($(wdots.south east)+(+0.1,-0.1)$);
	\draw[black,dashed] ($(x1.south west)+(-0.1,0.0)$) -- ($(zdots.south east)+(+0.1,-0.0)$);
	\draw [decorate,decoration={brace,amplitude=5pt,raise=4pt},yshift=0pt]
	($(zdots.north east)+(-0.0,0.0)$) -- ($(zdots.south east)+(-0.0,0.0)$) node [black,midway,xshift=0.6cm] {\footnotesize
		$d_1$};
	\draw [decorate,decoration={brace,amplitude=5pt,raise=4pt},yshift=0pt]
	($(wdots.north east)+(-0.0,0.0)$) -- ($(wdots.south east)+(-0.0,0.0)$) node [black,midway,xshift=0.6cm] {\footnotesize
		$d_2$};
	\end{tikzpicture}
	\caption{Overview of the structure of $\matrix{P}$ with axial \Emb s by \newcite{kitaev2020reformer}. They use two position embeddings, 
which can be interpreted as encoding a segment (bottom, $\matrix{P}^{(2)}$) and the position within that segment (top, $\matrix{P}^{(1)}$). This factorization is more parameter efficient, especially for long sequences.}	\figlabel{axial}
\end{figure}
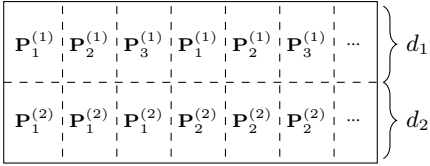

For very long sequences, i.e., large $\tmax{}$, the number of parameters added with $\matrix{P}$ is significant. Thus, \newcite{kitaev2020reformer} proposed a more parameter-efficient factorization called \name{axial position embeddings}. Although their method is not described in the paper, a description can be found in their code.\footnote{e.g., \url{https://huggingface.co/transformers/model_doc/reformer.html}} Intuitively, they have one embedding that marks a larger segment and a second embedding that indicates the position within each segment, see \figref{axial} for an overview. More specifically,  the matrix $\matrix{P}$ gets split into two embedding matrices $\matrix{P}^{(1)} \in \mathbb{R}^{t_1 \times d_1}$, $\matrix{P}^{(2)}  \in \mathbb{R}^{t_2 \times d_2}$ with $d=d_1 + d_2$ and $\tmax{} = t_1 t_2$. Then
\begin{equation}
\matrix{P}_{tj} = 
\begin{cases}
\matrix{P}^{(1)}_{r,j}\text{ if }j \leq d_1,\; r = t \bmod t_1\\
\matrix{P}^{(2)} _{s,j-d_1}\text{ if }j > d_1,\; s = \lfloor \frac{t}{t_1} \rfloor
\end{cases}
\end{equation}

\newcite{liu2020learning} argue that position embeddings should be parameter-efficient, data-driven, and should be able to handle sequences that are longer than any sequence in the training data. They propose a new model called \name{flow-based Transformer} or \name{FLOATER}, where they model position information with a continuous dynamic model. More specifically, consider $\matrix{P}$ as a sequence of timesteps $p_1, p_2, \dots, p_{\tmax{}}$. They suggest to model position information as a continuous function $p: \mathbb{R}_+ \to \mathbb{R}^d$ with 
\begin{equation}
p(t) = p(s) + \int_{s}^{t}h\left(\tau,  p(\tau), \theta_h\right) \diff\tau
\end{equation}
for $0 \leq s < t$ with some initial value for $p(0)$, where $h$ is some function, e.g., a neural network with parameters $\theta_h$. In the simplest case they then define $p_i := p(i \Delta t)$ for some fixed offset $\Delta t$. They experiment both with adding the information only in the first layer and at each layer (layerwise \ape). Even though they share parameters across layers, they use different initial values $p(0)$ and thus have different position embeddings at each layer. Sinusoidal position embeddings (cf.\secref{sinus}) are a special case of their dynamic model. Further, they provide a method to use the original position embeddings of a pretrained Transformer model while adding the dynamic model during finetuning only. In their experiments they observe that FLOATER outperforms learned and sinusoidal position embeddings, especially for long sequences. Further, adding position information at each layer increases performance.

Another approach to increase the Transformer efficiency both during training and inference is to keep $\tmax{}$ small.
The \name{Shortformer} by \newcite{press2020shortformer} caches previously computed unit representations
and therefore does not need to handle a large number of units at the same time.
This is made possible by what they call \name{position-infused attention},
where the position embeddings are added to the keys and queries, but not the values. Thus, the values are position independent and representations from previous subsequences can seamlessly be processed. More specifically, they propose
\begin{align}
	\matrix{\tilde{A}} \sim (\matrix{U} + \matrix{P}) \matrix{W}^{(q)} \tr{\matrix{W}^{(k)}}\tr{(\matrix{U} + \matrix{P})}\\
	\matrix{\tilde{M}} = \text{SoftMax} (\matrix{\tilde{A}}) \matrix{U} \matrix{W}^{(v)} \nonumber
\end{align}
The computation of the attention matrix $\matrix{\bar{A}}$ still depends on absolute \encoding s in Shortformer,
but $\matrix{\bar{M}}$ does not contain it,
as it is only a weighted sum of unit embeddings in the first layer. 
Consequently, Shortformer can attend to outputs of previous subsequences
and the position information has to be added in each layer again.
\newcite{press2020shortformer} report large improvements in training speed,
as well as language modeling perplexity.

While the former approaches all follow the \ape{} methodology,
\newcite{wang2020encoding} propose an alternative to simply summing position and unit embeddings.
Instead of having one embedding per unit, they model the representation as a function over positions.
That is, instead of feeding $\matrix{U}_t + \matrix{P}_t$ to the model for position $t$,
they suggest to model the embedding of unit $u$ as a function $g^{(u)}: \mathbb{N} \to \mathbb{R}^d$ such that the unit has a different embedding depending on the position at which it occurs.
After having proposed desired properties for such functions (position-free offset and boundedness), they introduce \name{complex-valued unit embeddings} where their $k$-th component is defined as follows
\begin{equation}
g^{(u)}(t)_k = r^{(u)}_k \exp\left(\mathit{i} (\omega^{(u)}_k t+ \theta^{(u)}_k )\right)
\end{equation}
Then, $\vector{r}^{(u)}, \bm{\omega}^{(u)}, \bm{\theta}^{(u)} \in \mathbb{R}^d$ are learnable parameters that define the unit embedding for the unit $u$. Their approach can also be interpreted as having a word embedding, parameterized by $\vector{r}^{(u)}$, that is component-wise multiplied with a position embedding, parameterized by $\bm{\omega}^{(u)}, \bm{\theta}^{(u)}$. They test these position-sensitive unit embeddings not only on Transformer models, but also on static embeddings, LSTMs, and CNNs, and observe large improvements.

\subsubsection{Relative \Encoding s}
\seclabel{seq:rel}

Among the first, \newcite{shaw2018self} introduced an alternative method for incorporating both absolute and \name{relative position encodings}. In their absolute variant they propose to change the computation to
\begin{equation}\eqlabel{shaw}
\matrix{A}_{ts} \sim \eat{\sqrt{\frac{1}{d_h}}} \tr{\matrix{U}_t} \matrix{W}^{(q)} \left(\tr{\matrix{W}^{(k)}} \matrix{U}_s  + \vector{a}^{(k)}_{(t,s)} \right)
\end{equation}
where $\vector{a}^{(k)}_{(t,s)} \in \mathbb{R}^d$ models the interaction between positions $t$ and $s$. Further they modify the computation of the values to 
\begin{equation}
\matrix{M}_{t} = \sum_{s = 1}^{\tmax{}}\text{SoftMax} (\matrix{A})_{ts} \left( \tr{\matrix{W}^{(v)}}\matrix{U}_s + \vector{a}^{(v)}_{(t,s)}\right)
\end{equation}
where $\vector{a}^{(v)}_{(t,s)} \in \mathbb{R}^{d}$ models again the interaction. While it cannot directly be compared  with the effect of simple addition of position embeddings, they roughly omit the position-position interaction and have only one unit-position term. In addition, they do not share the projection matrices but directly model the pairwise position interaction with the vectors $ \vector{a}$. In an ablation analysis they found that solely adding $\vector{a}^{(k)}_{(t,s)}$ might be sufficient.

To achieve relative positions they simply set
\begin{equation}
\vector{a}^{(k)}_{(t,s)} := \vector{w}^{(k)}_{\left(\mathit{clip}(s - t, r)\right)},
\end{equation}
where $\mathit{clip}(x, r) = \max\left(-r, \min(r, x)\right)$
and $\vector{w}^{(k)}_{(t)}\in \mathbb{R}^d$ for $-r \leq t \leq r$ for a maximum relative distance $r$.
Analogously for $\vector{a}^{(v)}_{(t,s)}$.
To reduce space complexity, they share the parameters across attention heads.
 While it is not explicitly mentioned in their paper we understand that they add the position information in each layer but do not share the parameters. The authors find that relative position embeddings perform better in machine translation and the combination of absolute and relative embeddings does not improve the performance.

\newcite{dai-etal-2019-transformer} propose the \name{Transformer XL} model. The main objective is to cover long sequences and to overcome the constraint of having a fixed-length context. To this end they fuse Transformer models with recurrence. This requires special handling of position information and thus a new \pim{}. At each attention head they adjust the computation of the attention matrix to 
\begin{equation}
\matrix{A}_{ts} \sim \underbrace{\tr{\matrix{U}_t}  \matrix{W}^{(q)}\tr{\matrix{W}^{(k)}} \matrix{U}_s}_{\text{content-based addressing}}  +
\underbrace{\tr{\matrix{U}_t}\matrix{W}^{(q)} \tr{\matrix{V}^{(k)}}  \vector{R}_{t - s}}_{\text{content-dependent position bias}} + 
\underbrace{\tr{\vector{b}} \tr{\matrix{W}^{(k)}}  \matrix{U}_{s}}_{\text{global content bias}} +
\underbrace{\tr{\vector{c}} \tr{\matrix{V}^{(k)}} \matrix{R}_{t - s}}_{\text{global position bias}}, 
\end{equation}
where $\matrix{R} \in \mathbb{R}^{\tau \times d}$ is a sinusoidal position embedding matrix as in \cite{vaswani2017attention} and $\vector{b}, \vector{c} \in \mathbb{R}^d$ are learnable parameters. They use different projection matrices for the relative positions, namely $\matrix{V}^{(k)} \in \mathbb{R}^{d \times d}$. Note that Transformer-XL is unidirectional and thus $\tau = t_{m} + \tmax{} - 1$, where $t_{m}$ is the memory length in the model. Furthermore they add this mechanism to all attention heads and layers, while sharing the position parameters across layers and heads.

\begin{figure}
	\centering
	\includegraphics[width=0.9\linewidth]{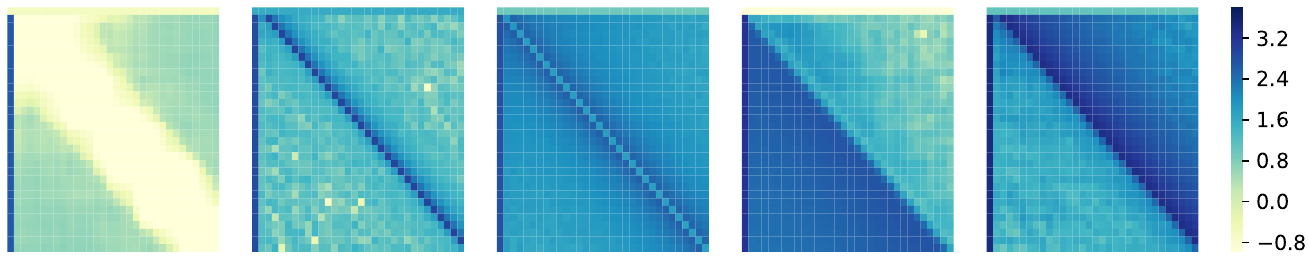}
	\caption{Figure by \newcite{ke2020rethinking}. Their position bias is independent of the input and can thus be easily visualized. The absolute position biases learn intuitive patterns as shown above. Patterns (from left to right) include ignoring position information, attending locally, globally, to the left, and to the right. One can clearly see the untied position bias for the first token, i.e., the [CLS] token, on the left and top of each matrix. \figlabel{ke2020}}
\end{figure}
There are more approaches that explore variants of \eqref{posfact}.
\newcite{ke2020rethinking} propose \name{untied position embeddings}. More specifically, they simply put $\matrix{U}$ into the Transformer and then modify the attention matrix $\matrix{A}$ in the first layer by adding a \emph{position bias}\begin{equation}
\matrix{A} \sim \matrix{U} \matrix{W}^{(q)} \tr{\matrix{W}^{(k)}} \tr{\matrix{U}} + \matrix{P} \matrix{V}^{(q)} \tr{\matrix{V}^{(k)} } \tr{\matrix{P}}
\end{equation} Compared to \eqref{posfact} they omit the unit-position interaction terms and use different projection matrices, $\matrix{V}^{(q)}, \matrix{V}^{(k)} \in \mathbb{R}^{d \times d}$ for units and positions.
Similarly, they add relative position embeddings by adding a scalar value. They add a matrix $\matrix{A}^r \in \mathbb{R}^{\tmax{} \times \tmax{}}$, where $\matrix{A}^r_{t, s} = \vector{b}_{t-s + \tmax{}}$ and $\vector{b} \in \mathbb{R}^{2\tmax{}}$ are learnable parameters, which is why we categorize this approach as \mam. A very similar idea with relative position encodings has also been used by \newcite{raffel2020exploring}.
\newcite{ke2020rethinking} further argue that the [CLS] token has a special role and thus they replace the terms $\tr{\matrix{P}_1} \matrix{V}^{(q)} \tr{\matrix{V}^{(k)}}\matrix{P}_s $ with a single parameter $\theta_1$ and analogously $\tr{\matrix{P}_t} \matrix{V}^{(q)} \tr{\matrix{V}^{(k)}}\matrix{P}_1 $ with $\theta_2$, i.e., they disentangle the position of the [CLS] token from the other position interactions.
They provide theoretical arguments that their absolute and relative position embeddings are complementary. Indeed, in their experiments the combination of relative and absolute embeddings boosts performance on the GLUE benchmark. They provide an analysis of the position biases learned by their network, see \figref{ke2020}.
A similar idea has been explored in \cite{dufter2020increasing},
where in a more limited setting, i.e., in the context of PoS-tagging,
learnable \emph{absolute or relative position biases} are learned instead of full position embeddings.

\newcite{chang-etal-2021-convolutions} provide a theoretical link between the \pim s proposed by \cite{shaw2018self,raffel2020exploring} and convolutions. They find that combining these two relative \pim{}s increases performance on natural language understanding tasks. 

Complementary to that line of research is a method by \newcite{he2020deberta}:
In their model \name{DeBERTa}, they omit the position-position interaction and focus on unit-position interactions. However, their embeddings are still untied or disentangled as they use different projection matrices for unit and position embeddings. They introduce relative position embeddings $\matrix{A}^r \in \mathbb{R}^{2\tmax{} \times d}$ and define 
\begin{equation}
\delta(t,s) = \left\{ \begin{array}{rcl}
0 & \mbox{if} & t-s \leq  -\tmax{}\\
2t_{\text{max}}-1 & \mbox{if} & t-s \geq \tmax{} \\
t-s+t_{\text{max}} & \mbox{else.}
\end{array}\right. 
\end{equation}
They then compute
\begin{align}
\matrix{A}_{ts} \sim \tr{\matrix{U}_t} \matrix{W}^{(q)} \tr{\matrix{W}^{(k)} } \matrix{U}_s + 
\tr{\matrix{U}_t} \matrix{W}^{(q)} \tr{\matrix{V}^{(k)}} \matrix{A}^r_{\delta(t,s)} + 
\tr{\matrix{A}^r_{\delta(s,t)} }\matrix{V}^{(q)} \tr{\matrix{W}^{(k)} } \matrix{U}_s
\end{align}
as the attention in each layer. While they share the weights of $\matrix{A}^r \in \mathbb{R}^{2\tmax{} \times d}$ across layers, the weight matrices are separate for each attention head and layer. In addition they change the scaling factor from $\sqrt{1 / d_h}$ to $\sqrt{1 / (3 d_h)}$. In the last layer they inject a traditional absolute position embedding matrix $\matrix{P}\in \mathbb{R}^{\tmax{}\times d}$. Thus they use both \mam{} and \ape{}.
They want relative encodings to be available in every layer but argue that the model should be reminded of absolute encodings right before the masked language model prediction.
In their example sentence \emph{a new store opened beside the new mall} they argue that \emph{store} and \emph{mall} have similar relative positions to \emph{new} and thus absolute positions are required for predicting masked units.

The following two approaches do not work with embeddings, but instead propose a direct multiplicative smoothing on the attention matrix and can thus be categorized as \mam.
\newcite{wu2020datransformer} propose a smoothing based on relative positions in their model \emph{DA-Transformer}. They consider the matrix of absolute values of relative distances $\matrix{R}\in\mathbb{N}^{\tmax{}\times \tmax{}}$ where $\matrix{R}_{ts} = |t - s|$. For each attention head $m$ they obtain $\matrix{R}^{(m)} = w^{(m)}\matrix{R}$ with $w^{(m)} \in \mathbb{R}$ being a learnable scalar parameter. They then compute 
\begin{equation}
\matrix{A} \sim \eat{\sqrt{\frac{1}{d_h}} }\text{ReLU}\left((\matrix{X} \matrix{W}^{(q)} \tr{\matrix{W}^{(k)}}\tr{\matrix{X}}) \circ\matrix{\hat{R}}^{(m)}\right),
\end{equation}
where $\matrix{\hat{R}}^{(m)}$ is a rescaled version of $\matrix{R}^{(m)}$ and $\circ$ is component-wise multiplication. For rescaling they use a learnable sigmoid function, i.e., 
\begin{equation}
	\matrix{\hat{R}}^{(m)} = \frac{1 + \exp(v^{(m)})}{1 + \exp(v^{(m)} - \matrix{R}^{(m)})}
\end{equation}
Overall, they only add $2h$ parameters as each head has two learnable parameters.
Intuitively, they want to allow each attention head to choose whether to attend to long range or short range dependencies. Note that their model is direction-agnostic. The authors observe improvements for text classification both over vanilla Transformer, relative position encodings by \cite{shaw2018self}, Transformer-XL \cite{dai-etal-2019-transformer} and TENER \cite{yan2019tener}.

Related to the DA-Transformer, \newcite{huang2020improve} review absolute and relative position embedding methods and propose four \pim s with relative position encodings: (1) Similar to \cite{wu2020datransformer} they scale the attention matrix by 
\begin{equation}
\matrix{A} \sim \eat{\sqrt{\frac{1}{d_h}}} (\matrix{X} \matrix{W}^{(q)} \tr{\matrix{W}^{(k)}}\tr{\matrix{X}})\circ \matrix{R},
\end{equation}
where $\matrix{R}_{ts} = \vector{r}_{|s - t|}$ and $ \vector{r}\in \mathbb{R}^{\tmax{}}$ is a learnable vector. (2) They consider $\matrix{R}_{ts} = \vector{r}_{s- t}$ as well to distinguish different directions. 
(3) As a new variant they propose 
\begin{equation}
\matrix{A}_{ts} \sim \eat{\sqrt{\frac{1}{d_h}}} \mathit{sum\_product}(\tr{\matrix{W}^{(q)}}\matrix{X}_t, \tr{\matrix{W}^{(k)}} \matrix{X}_s , \vector{r}_{s - t}),
\end{equation}
where $\vector{r}_{s - t} \in \mathbb{R}^{d}$ are learnable parameters and $\mathit{sum\_product}$ is the scalar product extended to three vectors. (4) Last, they extend the method by \newcite{shaw2018self} to not only add relative positions to the key, but also to the query in \eqref{shaw}, and in addition remove the position-position interaction. More specifically, 
\begin{equation}
\matrix{A}_{ts} \sim \eat{\sqrt{\frac{1}{d_h}}} \tr{\left(\tr{\matrix{W}^{(q)} } \matrix{U}_t  + \vector{r}_{s-t} \right)}\left(\tr{\matrix{W}^{(k)}} \matrix{U}_s  + \vector{r}_{s-t} \right) - \tr{\vector{r}_{s-t} }\vector{r}_{s-t} 
\end{equation}
 On several GLUE tasks \cite{wang-etal-2018-glue} they find that the last two methods perform best.

The next two approaches are not directly related to relative position encodings, but they can be interpreted as using relative position information. \newcite{shen2018disan} do not work directly with a Transformer model. Still they propose \name{Directional Self-Attention Networks (Di-SAN)}. Besides other differences to plain self-attention, e.g., multidimensional attention, they notably mask out the upper/lower triangular matrix or the diagonal in $\matrix{A}$ to achieve non-symmetric attention matrices.
Allowing attention only in a specific direction
does not add position information directly, but still makes the attention mechanism position-aware to some extent, i.e., enables the model to distinguish directions.

\begin{figure}
	\centering
	\includegraphics[width=0.75\linewidth]{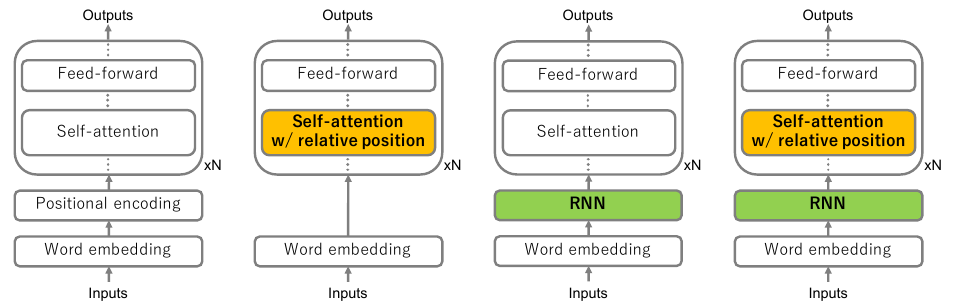}
	\caption{Figure by \newcite{neishi2019relation}. Overview of the architecture when using an RNN for learning position information. They combine their RNN-Transformer with relative position embeddings by \newcite{shaw2018self} in a model called \name{RR-Transformer} (far right). \figlabel{neishi2019}}
\end{figure}
\newcite{neishi2019relation} argue that recurrent neural networks (RNN) in form of gated recurrent units (GRU) \cite{cho2014learning} are able to encode relative positions. Thus they propose to replace position encodings by adding a single GRU layer on the input before feeding it to the Transformer, see \figref{neishi2019}. With their models called \name{RRN-Transformer} they observe comparable performance compared to position embeddings, however for longer sequences the GRU yields better performance. Combining their approach with the method by \newcite{shaw2018self} improves performance further, a method they call \name{RR-Transformer}.

Relative \pim{}s usually require the computation of the full attention matrix $\matrix{A}$
because each cell depends on a different kind of relative position interaction.
\citet{liutkus-etal-2021-relative} proposed an alternative called \emph{Stochastic Positional Encoding (SPE)}.
By approximating relative position interactions as cross-covariance structures of correlated Gaussian processes,
they make relative \encoding{}s available to linear-complexity Transformers, such as the Performer \citep{choromanski-etal-2021-rethinking},
that do not compute the full attention matrix, which would lead to a quadratic complexity.
Notably, \citet{liutkus-etal-2021-relative} also propose a gating mechanism
that controls with a learnable parameter how much the attention in each vector dimension depends on content vs.\ position information. 
They report performance improvements using SPE compared to absolute \encoding{}s for tasks involving long-range dependencies \citep{tay-etal-2021-long}.

\subsection{Sinusoidal}
\seclabel{sinus}

Another line of work experiments with sinusoidal values that are kept fixed during training to encode position information in a sequence.
The approach proposed by \newcite{vaswani2017attention}
is an instance of the absolute position \ape{} pattern,
called \name{sinusoidal position embeddings}, defined as
\begin{equation}
\matrix{P}_{tj} = 
\begin{cases}
\sin(10000^{-\frac{j}{d}}t)\text{ if }j\text{ even}\\
\cos(10000^{-\frac{(j - 1)}{d}}t)\text{ if }j\text{ odd}
\end{cases}
\end{equation}
They observe comparable performance between learned absolute position embeddings and their sinusoidal variant. However, they hypothesize that the sinusoidal structure helps for long range dependencies. This is for example verified by \newcite{liu2020learning}.
An obvious advantage is also that they can handle sequences of arbitrary length, which most position models cannot.
They are usually kept fixed and are not changed during training and thus very parameter-efficient.

\begin{figure}
	\centering
	\includegraphics[width=0.5\linewidth]{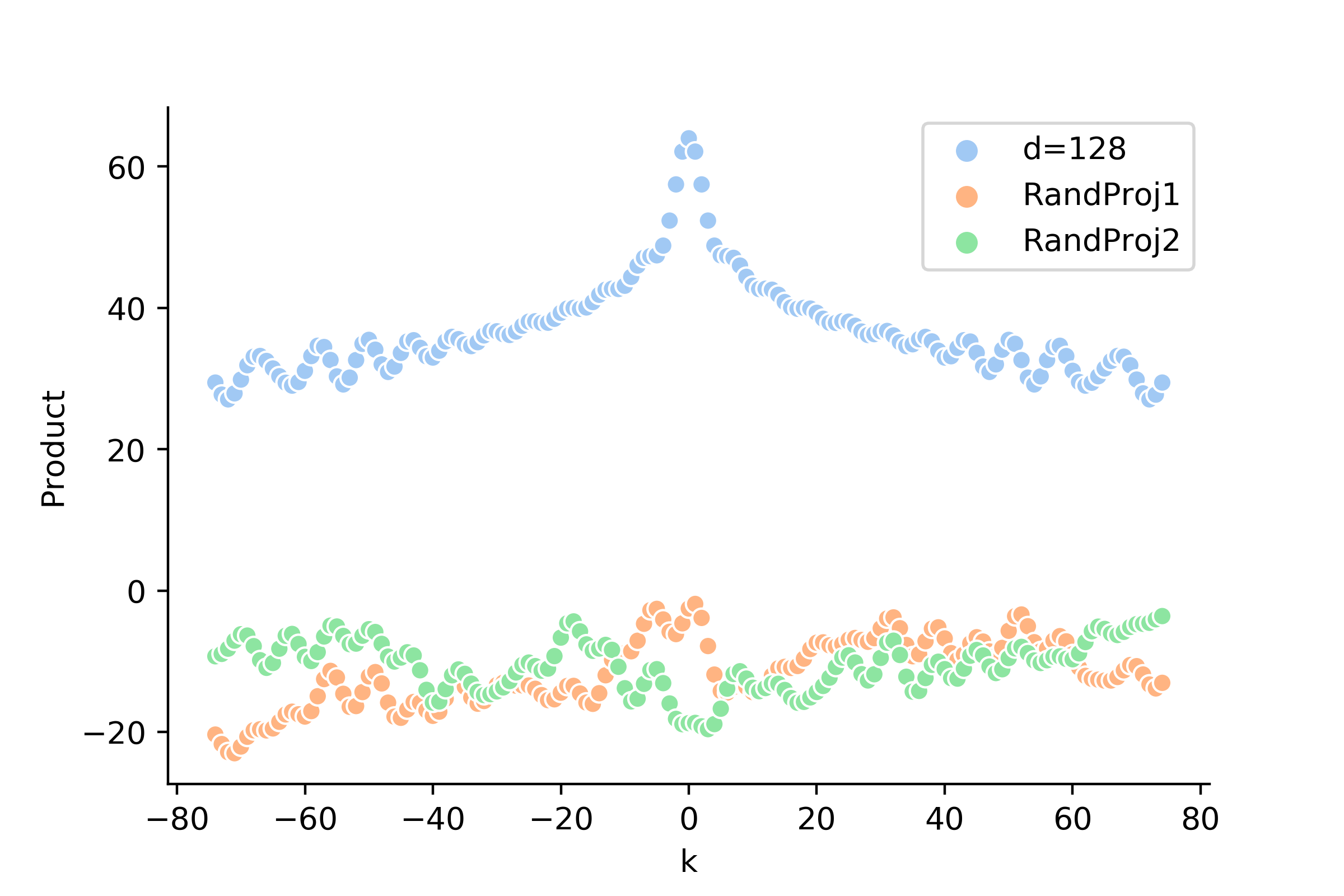}
	\caption{Figure by \newcite{yan2019tener}. Shown is the value of dot product on the y-axis between sinusoidal position embeddings with different relative distance ($k$) shown on the x-axis. The blue line shows the dot product without projection matrices and the other two lines with random projections. Relative position without directionality can be encoded without projection matrices, but with the projections this information is destroyed. \figlabel{yan2019}}
\end{figure}

Indeed, sinusoidal position embeddings exhibit useful properties in theory.  \newcite{yan2019tener} investigate the dot product of sinusoidal position embeddings and prove important properties:
\begin{enumerate*}[label=(\arabic*)]
	\item The dot product of two sinusoidal position embeddings depends only on their relative distance. That is, 
	$\tr{\matrix{P}_{t}}\matrix{P}_{t + r}$ is independent of $t$.
	\item $\tr{\matrix{P}_{t}}\matrix{P}_{t  - r} = \tr{\matrix{P}_{t}}\matrix{P}_{t + r}$, which means that sinusoidal position embeddings are unaware of direction.
\end{enumerate*}
However, in practice the sinusoidal embeddings are projected with two different projection matrices, which destroys these properties, see \figref{yan2019}. Thus, \newcite{yan2019tener}  propose a \name{Direction- and Distance-aware Attention} in their model \name{TENER} that maintains these properties and can, in addition, distinguish between directions. They compute
\begin{align}
\matrix{A}_{ts} \sim 
\underbrace{\tr{\matrix{U}_t} \matrix{W}^{(q)} \tr{\matrix{W}^{(k)} } \matrix{U}_s}_{\text{unit-unit}} + 
\underbrace{\tr{\matrix{U}_t} \matrix{W}^{(q)} \matrix{R}_{t - s}}_{\text{unit-relative position}} + 
\underbrace{\tr{\vector{u}} \tr{\matrix{W}^{(k)} } \matrix{U}_s}_{\text{unit-bias}} + 
\underbrace{\tr{\vector{v}} \matrix{R}_{t - s}}_{\text{relative position bias}}, 
\end{align}
where $\matrix{R}_{t - s} \in \mathbb{R}^d$ is a sinusoidal relative position vector defined as 
 \begin{align}
\matrix{R}_{t - s,j} =
\begin{cases}
\sin((t - s)10000^{-\frac{j}{d}})\text{ if }j\text{ even}\\
\cos((t - s)10000^{-\frac{(j - 1)}{d}})\text{ if }j\text{ odd,}
\end{cases}
\end{align}
and $\vector{u}, \vector{v} \in \mathbb{R}^d$ are learnable parameters for each head and layer.
In addition, they set $\matrix{W}^{(k)}$ to the identity matrix and omit the scaling factor $1/\sqrt{d}$ as they find that it performs better. Overall, the authors find massive performance increases for named entity recognition compared to standard Transformer models.
 
 \newcite{dehghani2019universal} use a variant of sinusoidal position embeddings in their \name{Universal Transformer}. In their model they combine Transformers with the recurrent inductive bias of recurrent neural networks. The basic idea is to replace the layers of a Transformer model with a single layer that is recurrently applied to the input, that is they share the weights across layers. In addition they propose conditional computation where they can halt or continue computation for each position individually. 
 When $l$ denotes their $l$-th application of the Transformer layer to the input, they add the position embeddings as follows
 \begin{align}
 \matrix{P}^l_{t, j} =
 \begin{cases}
 \sin(10000^{- \frac{j}{d}} t) + \sin(10000^{-\frac{j}{d}} l)\text{ if }j\text{ even}\\
 \cos(10000^{-\frac{j-1}{d}} t) + \cos(10000^{-\frac{j-1}{d}} l)\text{ if }j\text{ odd}
 \end{cases}
 \end{align}
 Their approach can be interpreted as adding sinusoidal position embeddings at each layer.

\newcite{li2019augmented} argue that the variance of sinusoidal position embeddings per position across dimensions varies greatly: for small positions it is rather small and for large positions it is rather high. The authors consider this a harmful property and propose \name{maximum variances position embeddings (mvPE)} as a remedy. They change the computation to 
\begin{equation}
\matrix{P}_{tj} = 
\begin{cases}
\sin(10000^{-\frac{j}{d}}kt)\text{ if }j\text{ even}\\
\cos(10000^{-\frac{j - 1}{d}}kt)\text{ if }j\text{ odd}\\
\end{cases}
\end{equation}
They claim that suitable values for the hyperparameter $k$ are $k>1000$.

\newcite{likhomanenko2021cape} introduce \name{continuous
  augmented positional embeddings} and focus to make
sinusoidal position embeddings work for other modalities
such as vision or speech. More specifically, they propose to
convert discrete positions to a continuous range and suggest noise augmentations to avoid that the model takes up spurious correlations. Instead of using the position $t$ in sinusoidal position embeddings they create $t'$ using mean normalization followed by a series of three random augmentations: 
\begin{enumerate*}[label=(\arabic*)] \item global shift $t' = t + \Delta$, \item local shift $t'=t + \epsilon$, \item global scaling $t'=\lambda t$. \end{enumerate*} $\Delta \sim \mathcal{U}(-\Delta_{\max}, \Delta_{\max})$, $\epsilon \sim \mathcal{U}(-\epsilon_{\max}, \epsilon_{\max})$, and $\lambda \sim \mathcal{U}(-\log(\lambda_{\max}), \log(\lambda_{\max}))$ are sampled from a uniform distribution. Note that during inference only mean normalization is performed. As expected, they find their model to work well on vision and speech data. On natural language it performed on par with minor improvements compared to sinusoidal position embeddings as measured on machine translation.

\newcite{su2021roformer} propose to multiply sinusoidal position embeddings rather than adding them in their model \name{rotary position embeddings}. Intuitively, they rotate unit representations according to their position in a sequence. More specifically, they modify the attention computation to 
\begin{equation}
	\matrix{A}_{ts} \sim \tr{\matrix{U}_t} \matrix{W}^{(q)} \matrix{R}_{\Theta, t - s}\tr{\matrix{W}^{(k)} } \matrix{U}_s
\end{equation} where $\matrix{R}_{\Theta, t - s} = \tr{\matrix{R}_{\Theta, s}}  \matrix{R}_{\Theta, t}$ with  $\matrix{R}_{\Theta, s} \in \mathbb{R}^{d \times d}$ is a block-diagonal matrix with rotation matrices on its diagonal. They only provide results on Chinese data where they are able to match the performance of learned absolute position embeddings and claim that their approach is beneficial for long sequences.

\subsection{Graphs}

In the following section,
we will take a look at \pim s for graphs,
i.e., cases where
Transformers have been used for genuine graph input
as well as cases where the graph is used as a sentence representation,
e.g., a dependency graph.
We distinguish two types of graph position models
according to the assumptions they make about the graph structure:
positions in hierarchies (trees) and arbitrary graphs.

\subsubsection{Hierarchies (Trees)}
\seclabel{trees}

\begin{figure}
	\centering
	\includegraphics[width=0.7\linewidth]{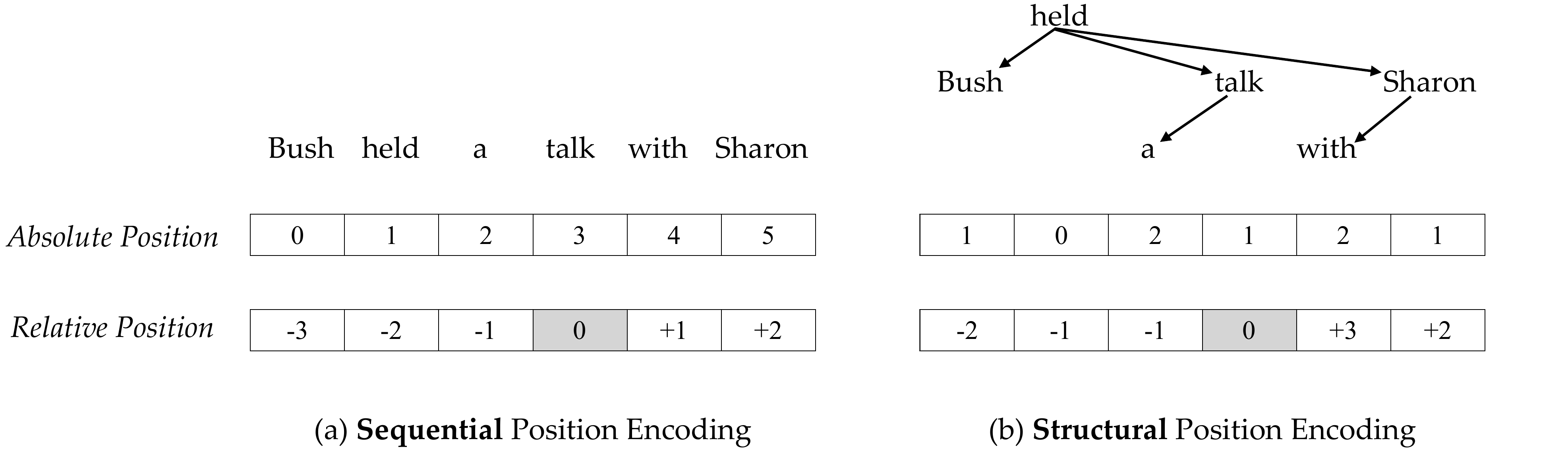}
	\caption{Figure by \newcite{wang2019self}. They compute absolute and relative encodings not based on the sequential order of a sentence (left), but based on a dependency tree (right). Both absolute and relative encodings can be created.}
	\figlabel{wang2019}
\end{figure}

\newcite{wang2019self} propose \name{structural position representations} (SPR), see \figref{wang2019}.
This means that instead of treating a sentence as a sequence of information,
they perform dependency parsing and compute distances on the parse tree (dependency graph).\footnote{Dependency parsers usually do not operate on subwords. So subwords are assigned the position of their main word.}
We can distinguish two settings:
\begin{enumerate*}[label=(\arabic*)]
	\item Analogously to absolute position encodings in sequences,
		where unit $u_t$ is assigned position $t$,
		absolute SPR assigns $u_t$ the position $\mathit{abs}(u_t) := d_{\text{tree}}(u_t, \textsc{root})$
		where \textsc{root} is the root of the dependency tree, i.e., the main verb of the sentence, 
		and $d_{\text{tree}}(x, y)$ is the path length between $x$ and $y$ in the tree.
	\item For the relative SPR between the units $u_t, u_s$, they define $\mathit{rel}(u_t, u_s) = \mathit{abs}(u_t) - \mathit{abs}(u_s)$ if $u_t$ is on the path from $u_s$ to the root or vice versa. Otherwise, they use 
	$\mathit{rel}(u_t, u_s) = \mathit{sgn}(t - s) (\mathit{abs}(u_t) + \mathit{abs}(u_s))$.
\end{enumerate*}
So we see that SPR does not only assume the presence of a graph hierarchy
but also needs a strict order to be defined on the graph nodes,
because $\mathit{rel}$ equally encodes sequential relative position.
This makes SPR a suitable choice for working with dependency graphs
but renders SPR incompatible with other tree structures. 

Having defined the position of a node in a tree,
\citet{wang2019self} inject their SPR via sinusoidal \ape{} for absolute and
via learned embeddings in combination with \mam{} for relative positions.
It is noteworthy, that \citet{wang2019self} achieve their best performance by combining both variants of SPR with sequential position information
and that SPR as sole sentence representation,
i.e., without additional sequential information, leads to a large drop in performance.

\newcite{shiv2019novel} propose alternative absolute \name{tree position encodings} (TPE).
They draw inspiration from the mathematical properties of sinusoidals but do not use them directly like \citet{wang2019self}. 
Also unlike SPR, their position encodings consider the full path from a node to the root of the tree and not only its length, thus assigning every node a unique position. This is more in line with the spirit of absolute sequential position models (\secref{seq:abs}).
The first version of TPE is parameter-free:
The path from the root of an $n$-ary tree to some node is defined as the individual decisions that lead to the destination,
i.e., which of the $n$ children is the next to be visited at each intermediate step.
These decisions are encoded as one-hot vectors of size $n$.
The whole path is simply the concatenation of these vectors (padded with 0s for shorter paths).
In a second version, multiple instances of parameter-free TPE are concatenated
and each one is weighted with a different learned parameter.
After scaling and normalizing these vectors, they are added to the unit embeddings before the first Transformer layer (\ape).

\subsubsection{Arbitrary Graphs}
\seclabel{graphs}

\begin{figure}[t]
	\centering
	\begin{subfigure}{.37\linewidth}
		\raggedright
		\includegraphics[page=3,width=\linewidth]{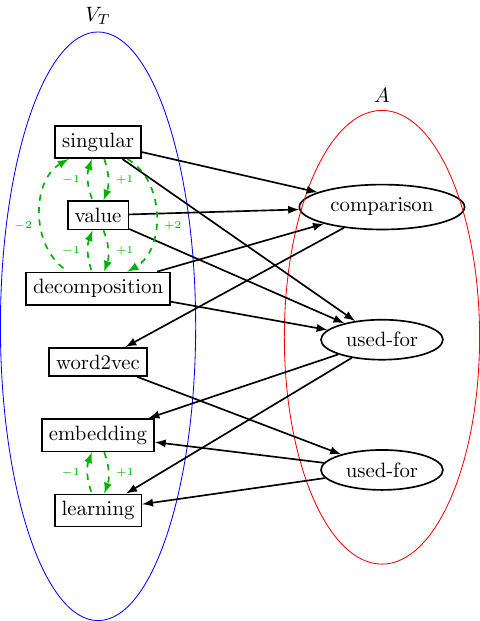}
	\end{subfigure}
	\hfill
	\begin{subfigure}{.53\linewidth}
	\small
	\raggedleft
	\def\intercolumn{\hspace{.4em}}
	\def\outspace{\hspace{.4em}}
	\def\cellwidth{1.25em}
	\newcolumntype{x}[1]{>{\centering\arraybackslash\hspace{0pt}}p{#1}}
	\begingroup
	\renewcommand*{\arraystretch}{1.1}
	\begin{tabular}{@{\outspace}x{\cellwidth}@{\intercolumn}|@{\intercolumn}x{\cellwidth}@{\intercolumn}|@{\intercolumn}x{\cellwidth}@{\intercolumn}|@{\intercolumn}x{\cellwidth}@{\intercolumn}|@{\intercolumn}x{\cellwidth}@{\intercolumn}|@{\intercolumn}x{\cellwidth}@{\intercolumn}|@{\intercolumn}x{\cellwidth}@{\intercolumn}|@{\intercolumn}x{\cellwidth}@{\intercolumn}|@{\intercolumn}x{\cellwidth}@{\intercolumn}|@{\intercolumn}x{\cellwidth}@{\outspace}}
		&\multicolumn{6}{@{\intercolumn}c@{\intercolumn}|@{\intercolumn}}{$V_T$} & \multicolumn{3}{@{\intercolumn}c@{\intercolumn}}{$A$}\\[.25em]
		\hline
		& \texttt{s} & \texttt{v} & \texttt{d} & \texttt{w} & \texttt{e} & \texttt{l} & \texttt{c} & \texttt{u1} & \texttt{u2} \\
		\hline
		\texttt{s} & 0 & 4 & 5 & 2 & 2 & 2 & 1 & 1 & 3 \\
		\hline
		\texttt{v} & -4 & 0 & 4 & 2 & 2 & 2 & 1 & 1 & 3 \\
		\hline
		\texttt{d} & -5 & -4 & 0 & 2 & 2 & 2 & 1 & 1 & 3 \\
		\hline
		\texttt{w} & -2 & -2 & -2 & 0 & 2 & 2 & -1 & $\infty$ & 1 \\
		\hline
		\texttt{e} & -2 & -2 & -2 & -2 & 0 & 4 & -3 & -1 & -1 \\
		\hline
		\texttt{l} & -2 & -2 & -2 & -2 & -4 & 0 & -3 & -1 & -1 \\
		\hline
		\texttt{c} & -1 & -1 & -1 & 1 & 3 & 3 & 0 & $\infty$ & 2 \\
		\hline
		\texttt{u1} & -1 & -1 & -1 & $\infty$ & 1 & 1 & $\infty$ & 0 & $\infty$ \\
		\hline
		\texttt{u2} & -3 & -3 & -3 & -1 & 1 & 1 & -2 & $\infty$ & 0 \\
		\hline
	\end{tabular}
	\endgroup
	\end{subfigure}
	
	\caption{Figure from \citep{schmitt2021modeling}, showing their definition of relative position encodings in a graph based on the lengths of shortest paths. $\infty$ means that there is no path between two nodes. Numbers higher than $3$ and lower than $-3$ represent sequential relative position in multi-token node labels (dashed green arrows).}
	\figlabel{graformer}
\end{figure}
 
\newcite{zhu-etal-2019-modeling} were the first to propose a Transformer model
capable of processing arbitrary graphs.
Their \pim{} solely defines the relative position between nodes
and incorporates this information by manipulating the attention matrix (\mam):
\begin{align}
	\matrix{A}_{ts} &\sim \eat{\sqrt{\frac{1}{d_h}}} \tr{\matrix{U}_t} \matrix{W}^{(q)} \left(\tr{\matrix{W}^{(k)}} \matrix{U}_s  + \tr{\matrix{W}^{(r)}} \vector{r}_{(t,s)} \right)\\
	\matrix{M}_{t} &= \sum_{s = 1}^{\tmax{}}\text{SoftMax} (\matrix{A})_{ts} \left( \tr{\matrix{W}^{(v)}}  \matrix{U}_s +  \tr{\matrix{W}^{(f)}} \vector{r}_{(t,s)} \right) \nonumber
\end{align}
where $\matrix{W}^{(r)}, \matrix{W}^{(f)}\in\mathbb{R}^{d\times d}$ are additional learnable parameters,
and $\vector{r}_{(t,s)} \in \mathbb{R}^d$ is a representation of the sequence of edge labels
and special edge direction symbols ($\uparrow$ and $\downarrow$) on the shortest path between the nodes $u_t$ and $u_s$.
\newcite{zhu-etal-2019-modeling} experiment with 5 different ways of computing $\vector{r}$,
where the best performance is achieved by two approaches:
\begin{enumerate*}[label=(\arabic*)]
	\item A CNN with $d$ kernels of size 4 that convolutes the embedded label sequence $\matrix{U}^{(r)}$ into $\vector{r}$ (cf.\ \citealp{kalchbrenner-etal-2014-convolutional}) and
	\item a one-layer self-attention module with sinusoidal position embeddings $\matrix{P}$ (cf.\ \secref{sinus}):
\end{enumerate*}
\begin{align}
	\matrix{A}^{(r)} &\sim \eat{\sqrt{\frac{1}{d}}} (\matrix{U}^{(r)} + \matrix{P}) \matrix{W}^{(qr)} \tr{\matrix{W}^{(kr)}}\tr{(\matrix{U}^{(r)} + \matrix{P})}\nonumber\\
	\matrix{M}^{(r)} &= \text{SoftMax} (\matrix{A}^{(r)}) (\matrix{U}^{(r)} + \matrix{P})\matrix{W}^{(vr)}\\
	\vector{a}^{(r)} &= \text{SoftMax} (\matrix{W}^{(r_2)} \mathit{tanh} (\matrix{W}^{(r_1)} \tr{\matrix{M}^{(r)}} ) )\nonumber\\
	\vector{r} &= \sum_{k=1}^{\tmax{}^{(r)}} a^{(r)}_k \matrix{M}^{(r)}_k \nonumber
\end{align}
with $\matrix{W}^{(r_1)}\in\mathbb{R}^{d_r \times d}, \matrix{W}^{(r_2)}\in\mathbb{R}^{1\times d_r}$ additional model parameters.
While there is a special symbol for the empty path from one node to itself,
this method implicitly assumes that there is always at least one path between any two nodes.
While it is easily possible to extend this work to disconnected graphs by introducing another special symbol,
the effect on performance is unclear.

\newcite{cai-lam-2020-graph} also define relative position in a graph based on shortest paths.
They differ from the former approach in omitting the edge direction symbols
and using a bidirectional GRU \citep{cho2014learning}, to aggregate the label information on the paths (cf.\ the RNN-Transformer described by \newcite{neishi2019relation}).
After linearly transforming the GRU output,
it is split into a forward and a backward part: $[\vector{r}_{t\to s}; \vector{r}_{s\to t}] = \matrix{W}^{(r)} \mathit{GRU}(\dots)$.
These vectors are injected into the model in a variant of \ape{}\begin{align}
	\matrix{A}_{st} &\sim\tr{ (\matrix{U}_s + \vector{r}_{s\to t}) }\matrix{W}^{(q)} \tr{\matrix{W}^{(k)}} (\matrix{U}_t + \vector{r}_{t\to s}) \nonumber\\
	&= \underbrace{\tr{\matrix{U}_s } \matrix{W}^{(q)}\tr{\matrix{W}^{(k)}} \matrix{U}_t}_{\text{content-based addressing}}  +
	\underbrace{\tr{\matrix{U}_s}\matrix{W}^{(q)} \tr{\matrix{W}^{(k)}}  \vector{r}_{t\to s}}_{\text{source relation bias}} \\
	&\phantom{{}=} + 
	\underbrace{\tr{\vector{r}_{s\to t} }\matrix{W}^{(q)} \tr{\matrix{W}^{(k)}}  \matrix{U}_{t}}_{\text{target relation bias}} +
	\underbrace{\tr{\vector{r}_{s\to t}} \matrix{W}^{(q)} \tr{\matrix{W}^{(k)}} \vector{r}_{t\to s}}_{\text{universal relation bias}}\nonumber
\end{align} It is noteworthy that \citet{cai-lam-2020-graph} additionally include absolute SPR (see \secref{trees}) in their model to exploit the hierarchical structure of the abstract meaning representation (AMR) graphs they evaluate on.
It is unclear which position model has more impact on performance. 

\citet{schmitt2021modeling} avoid computational overhead in their \name{Graformer} model
by defining relative position encodings in a graph as the length of shortest paths instead of the sequence of edge labels (see \figref{graformer} for an example):
\begin{equation}
	r_{(t, s)} =
	\begin{cases}
		\infty, & \text{if there is no path between $t, s$} \\
		\begin{aligned}
			&\text{sequential relative position of $u_t, u_s$}\\
			&\text{shifted by a constant to avoid clashes},
		\end{aligned} & \text{if subwords $u_t, u_s$ from the same word} \\
		d_{\text{graph}}(t,s), & \text{if } d_{\text{graph}}(t, s) \leq d_{\text{graph}}(s, t) \\
		-d_{\text{graph}}(s, t), & \text{if } d_{\text{graph}}(t, s) > d_{\text{graph}}(s, t)
	\end{cases}
\end{equation}
where $d_{\text{graph}}(x,y)$ is the length of the shortest path between $x$ and $y$.
This definition also avoids the otherwise problematic case where there is more than one shortest path between two nodes
because the length is always the same even if the label sequences are not.
The so-defined position information is injected via learnable scalar embeddings as \mam{} similar to \cite{raffel2020exploring}.

In contrast to the other approaches, Graformer explicitly models disconnected graphs ($\infty$)
and does not add any sequential position information.
Unfortunately, \citet{schmitt2021modeling} do not evaluate Graformer on the same tasks as the other discussed approaches,
which makes a performance comparison difficult.

All the approaches discussed so far have in common that they allow any node to compute attention over the complete set of nodes in the graph -- similar to the global self-attention over tokens in the original Transformer --
and that they inject the graph structure solely over a relative \encoding{}.
\citet{dwivedi2021generalization} restrict attention in their graph Transformer to the local node neighborhood
and therefore do not need to capture the graph structure by defining the relative position between nodes.
Instead they employ an absolute \ape{} model by adding Laplacian Eigenvectors to the node embeddings before feeding them to the Transformer encoder.
Like sinusoidal \emb{}s only depend on the (unchanging) order of words,
Laplacian Eigenvectors only depend on the (unchanging) graph structure.
Thus, these \emb{}s are parameter-free and can be precomputed for efficient processing.
Again, however, an empirical comparison is impossible because \citet{dwivedi2021generalization} evaluate their model on node classification and graph regression
whereas the approaches discussed above are tested on graph-to-text generation.

A similarly parameter-free approach is described by \citet{zhang-etal-2020-graph}.
In their pretraining based on linkless subgraph batching,
they combine different features of each node,
both predefined (such as node labels) and structural information (such as shortest path lengths),
translate them to integers (the \encoding{}) and, finally, map them to real numbers via sinusoidal \emb{}s (see \secref{sinus}).
The final \textsc{graph-bert} model takes the overall sum as its input (\ape{}).

\subsection{Decoder}
\seclabel{decoder}

\newcite{takase-okazaki-2019-positional} propose a simple extension to sinusoidal embeddings by incorporating sentence lengths in the position encodings of the decoder. Their motivation is to be able to control the output length during decoding and to enable the decoder to generate any sequence length independent of what lengths have been observed during training. The proposed \name{length-difference position embeddings} are 
\begin{equation}
\matrix{P}_{tj} = 
\begin{cases}
\sin(10000^{-\frac{j}{d}}(l - t))\text{ if }j\text{ even}\\
\cos(10000^{-\frac{(j - 1)}{d}}(l - t))\text{ if }j\text{ odd}\\
\end{cases}
\end{equation}
where $l$ is a given length constraint. Similarly, they propose a \name{length-ratio position embedding} given by 
\begin{equation}
\matrix{P}_{tj} = 
\begin{cases}
\sin(l^{-\frac{j}{d}}t)\text{ if }j\text{ even}\\
\cos(l^{-\frac{(j - 1)}{d}}t)\text{ if }j\text{ odd}\\
\end{cases}
\end{equation}
The length constraint $l$ is the output length of the gold standard. They observe that they can control the output length effectively during decoding.
\newcite{oka-etal-2020-incorporating} extended this approach by adding noise to the length constraint (adding a randomly sampled integer to the length) and by predicting the target sentence length using the Transformer model. Although, in theory, these approaches could also be used in the encoder, above work focuses on the decoder.

\begin{figure}
	\centering
	\includegraphics[width=0.4\linewidth]{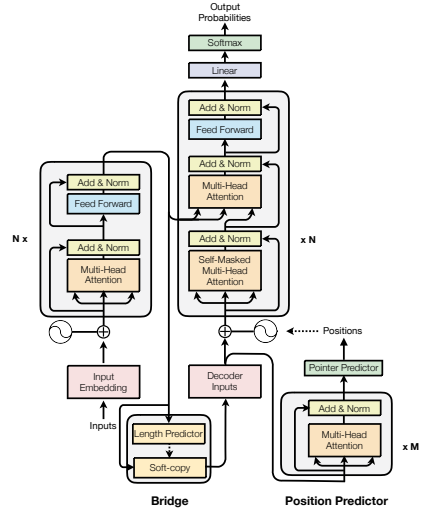}
	\caption{Figure by \newcite{bao2019non}. Overview of their PNAT architecture with the position prediction module. They use the encoder output to predict the output length and use a modified version as input to the decoder. The position predictor then predicts a permutation of position encodings for the output sequence. \figlabel{bao2019}}
\end{figure}
\newcite{bao2019non} propose to predict positions word units in the decoder in order to allow for effective non-autoregressive decoding, see \figref{bao2019}. More specifically, they predict the target sentence length and a permutation from decoder inputs and subsequently reorder the position embeddings in the decoder according to the predicted permutation. Their model called \name{PNAT} achieves performance improvements in machine translation.

\subsection{Crosslingual}
\seclabel{crosslingual}
Unit order across different languages is quite
different. English uses a subject-verb-object ordering
(SVO), but all possible orderings of S, V and O have been
argued to occur in the world's languages. Also, whereas unit
ordering is rather fixed in English, it
varies considerably in other languages, e.g, 
in German. This raises the question whether it is useful to share position information across languages.

Per default, position embeddings are shared in multilingual models \cite{devlin-etal-2019-bert,conneau2020unsupervised}.
\newcite{artetxe2020crosslingual} observe mixed results with \name{language specific position embeddings} in the context of transferring monolingual models to multiple languages: for most languages it helps, but for some it seems harmful. They experimented with learned absolute position embeddings as proposed in \cite{devlin-etal-2019-bert}.

\newcite{ding2020self} use crosslingual position embeddings (\name{XL PE}): in the context of machine translation, they obtain reorderings of the source sentence and subsequently integrate both the original and reordered position encodings into the model and observe improvements on the machine translation task.

\newcite{liu2020improving} find that position information hinders zero-shot crosslingual transfer in the context of machine translation. They remove a residual connection in a middle layer to break the propagation of position information, and thereby achieve large improvements in zero-shot translation.

Similarly, \newcite{liu2020do} find that unit order information harms crosslingual transfer, e.g., in a zero-shot transfer setting. They reduce position information by a) removing the position embeddings, and replacing them with one dimensional convolutions, i.e., leveraging only local position information, b) randomly shuffling the unit order in the source language, and c) using position embeddings from a multilingual model and freezing them. Indeed they find that reducing order information with these three methods increases performance for crosslingual transfer.

\subsection{Analysis}
\seclabel{analysis}

There is a range of work comparing and analyzing \pim s. 
\newcite{rosendahl2019analysis} analyze them in the context of machine translation. They find similar performance for absolute and relative encodings, but relative encodings are superior for long sentences. In addition, they find that the number of learnable parameters can often be reduced without performance loss. 

\newcite{yang2019assessing} evaluate the ability of recovering the original word positions after shuffling some input words.
In a comparison of recurrent neural networks, Transformer models, and DiSAN (both with learned position embeddings),
they find that RNN and DiSAN achieve similar performance on the word reordering task, whereas Transformer is worse. However, when trained on machine translation Transformer performs best in the word reordering task.

\begin{figure}
	\centering
	\includegraphics[width=0.9\linewidth]{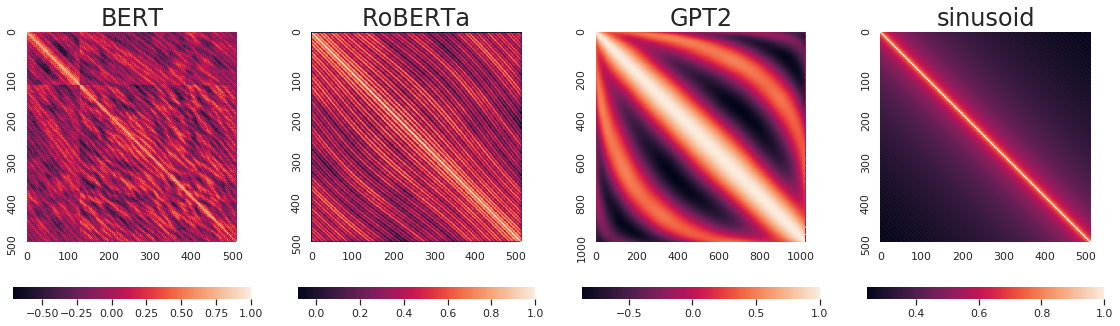}
	\caption{Figure by \newcite{wang2020what}. Shown is the position-wise cosine similarity of position embeddings (\ape{}) after pretraining. They compare three pretrained language models that use learned absolute position embeddings as in \cite{devlin-etal-2019-bert}, and sinusoidal positions as in \cite{vaswani2017attention}. BERT shows a cutoff at 128 as it is first trained on sequences with 128 tokens and subsequently extended to longer sequences. GPT-2 exhibits the most homogenous similarity patterns.}
	\figlabel{wang2020}
\end{figure}
\newcite{wang2020what} provide an in-depth analysis of what position embeddings in large pretrained language models learn. They compare the embeddings from BERT \cite{devlin-etal-2019-bert}, RoBERTa \cite{liu2019roberta}, GPT-2 \cite{radford2019language}, and sinusoidal embeddings. See \figref{wang2020} for their analysis.

More recently, \newcite{wang2021on} present an extensive analysis of position embeddings. They empirically compare 13 variants of position embeddings.
Among other findings, they conclude that absolute position embeddings are favorable for classification tasks
and relative embeddings perform better for span prediction tasks. 

\pagebreak

\input{content/comparison_table.tex}

\newcite{chen2021demystifying} compare absolute and relative position embeddings as introduced by \newcite{ke2020rethinking}. They slightly modify the formulation, add segment embeddings as used in the original BERT formulation \cite{devlin-etal-2019-bert} and investigate sharing parameters across heads and layers. They find that an argued superiority of relative position embeddings might have been due to the fact that they are added to each attention head. When applying the same procedure with absolute position embeddings they find the best performance across a range of natural language understanding tasks. 

We provide a high level comparison of the discussed methods in \tabref{comparison}. In this table we group similar approaches from a methodological point of view. The objective is to make comparisons easier and spot commonalities faster.

\section{Conclusion} 

We presented an overview of methods to inject position information into Transformer models.
We hope our unified notation and systematic comparison (\tabref{comparison}) will foster understanding and spark new ideas in this important research area. 

Open questions that we believe still need to be fully investigated
and would be promising starting points for future work include: 
\begin{enumerate}[label=(\arabic*)]
	\item How do current \pim s compare empirically
          on different tasks? Some analysis papers such as
          \citep{wang2021on} are extensive and provide
          many insights. Still, many aspects and differences
          of the \pim s are not fully understood. 
	\item How important is word order for specific tasks? For many tasks, treating sentences as bag-of-words could be sufficient. Indeed, \newcite{wang2021on} show that without position embeddings the performance drops for some tasks are marginal. Thus we consider it interesting to investigate for which tasks position information is essential. 
	\item Can we use \pim s to include more information
          about the structure of text? While there are many
          models for processing sequential and graph-based
          structures, there is a wide range of structural information in text that is not considered currently. Some examples include tables, document layout, list enumerations and sentence order. Could these structures be integrated with current \pim s or are new methods required for representing document structure?
\end{enumerate}

\section*{Acknowledgements}
This work was supported
by the European Research Council (\# 740516)
and by the BMBF as part of the project MLWin (01IS18050).
The first author was supported by the Bavarian research institute for digital transformation (bidt) through their fellowship program.
We also gratefully acknowledge a Ph.D.\ scholarship
awarded to the second author by the German Academic Scholarship Foundation (Studienstiftung des
deutschen Volkes). We thank Nikita Datcenko for helpful discussions and valuable insights.

%% file: content/comparison_table.tex
\begin{table}[H]
	\newcommand{\mr}[2]{\multirow{#1}{*}{#2}}
	\def\yes{\faCheck}
	\def\no{\faTimes}
	\def\inj{\faSyringe} 
	\def\lea{\faGraduationCap} 
	\def\rec{\faLayerGroup} 
	\def\sha{\faRecycle} 
	\def\posType{\faMapMarker*}  
	\def\unb{\faInfinity} 
	\def\param{\textbf{\#{}Param}}
	\def\other{\textbf{-}}
	\def\rotateangle{60}
	\scriptsize
	\renewcommand{\arraystretch}{1.5}
	\centering
	\def\mysep{0.05cm}
	\begin{tabular}{c@{\hspace{.2em}}l
			@{\hspace{\mysep}}c@{\hspace{\mysep}}
			@{\hspace{\mysep}}c@{\hspace{\mysep}}
			@{\hspace{\mysep}}c@{\hspace{\mysep}}
			@{\hspace{\mysep}}c@{\hspace{\mysep}}
			@{\hspace{\mysep}}c@{\hspace{\mysep}}
			@{\hspace{\mysep}}c@{\hspace{\mysep}}}
		\toprule
		& & \rotatebox[origin=lB]{\rotateangle}{Ref. Point}& \rotatebox[origin=lB]{\rotateangle}{Inject. Met.}& \rotatebox[origin=lB]{\rotateangle}{Learnable}& \rotatebox[origin=lB]{\rotateangle}{Recurring} & \rotatebox[origin=lB]{\rotateangle}{Unbound}& \\
		&\textbf{Model} & \posType{} & \inj{} & \lea{} & \rec{} & \unb{} & \param \\
		\cmidrule{2-8}
		\multirow{24}{0.5cm}{\rotatebox{90}{\textbf{Sequences}}}
		&Transformer w/ emb.\ \citep{vaswani2017attention}&\mr{3}{A}&\mr{3}{\ape}&\mr{3}{\yes}&\mr{3}{\no}&\mr{3}{\no}&\mr{2}{$\tmax{}d$}\\
		&BERT \citep{devlin-etal-2019-bert}&&&&&& \\
		&Reformer \citep{kitaev2020reformer}&&&&&&$(d-d_1)\frac{\tmax{}}{t_1} + d_1t_1$ \\	
		\cmidrule{2-8}
		&FLOATER \citep{liu2020learning}&A&\ape&\yes&\yes&\yes&$0$ or more\\
		&Shortformer \citep{press2020shortformer}&A&\ape&\no&\yes&\yes&0\\
		&\citet{wang2020encoding}&A&\other&\yes&\no&\yes&$3\tmax{}d$\\
		&\citet{shaw2018self} (abs)&A&\mr{1}{\mam}&\mr{1}{\yes}&\mr{1}{\yes}&\mr{1}{\no}&$2\tmax{}^2dl$\\
		\cmidrule{2-8}
		&\citet{shaw2018self} (rel)&\mr{3}{R}&\mr{3}{\mam}&\mr{3}{\yes}&\mr{3}{\yes}&\mr{3}{\no}&$2(2\tmax{}-1)dl$\\
		&T5 \citep{raffel2020exploring}&&&&&&$(2\tmax{}-1)h$\\
		&\citet{huang2020improve}&&&&&&$2dlh(\tmax{}-1)$\\
		\cmidrule{2-8}	
		&DeBERTa \citep{he2020deberta}&B&Both&\yes&\yes&\no&$\tmax{}d+2d^2$\\
		\cmidrule{2-8}	
		&Transformer XL \citep{dai-etal-2019-transformer}&\mr{2}{R}&\mr{2}{\mam}&\mr{2}{\yes}&\mr{2}{\yes}&\mr{2}{\yes}&$2d + d^2lh$\\
		&DA-Transformer \citep{wu2020datransformer}&&&&&&$2h$\\
		\cmidrule{2-8}	
		&TUPE \citep{ke2020rethinking}&\mr{2}{B}&\mr{2}{\mam}&\mr{2}{\yes}&\mr{2}{\no}&\mr{2}{\no}&$2d^2 + \tmax{} (d + 2)$\\
		&\citet{dufter2020increasing}&&&&&&$\tmax{}^2h + 2\tmax{}h$ \\
		\cmidrule{2-8}
		&RNN-Transf. \citep{neishi2019relation}&R&\other{}&\yes&\no&\yes&$6d^2+3d$\\
		\cmidrule{2-8}
		&SPE \citep{liutkus-etal-2021-relative}&R&\mam&\yes&\yes&\no&$3Kdh+ld$\\
		\cmidrule{2-8}
		&Transformer w/ sin.\ \citep{vaswani2017attention}&\mr{4}{A}&\mr{4}{\ape}&\mr{4}{\no}&\mr{4}{\no}&\mr{4}{\yes}&\mr{4}{$0$}\\
		&\citet{li2019augmented}&&&&&& \\
		&\citet{takase-okazaki-2019-positional}&&&&&&\\
		&\citet{oka-etal-2020-incorporating}&&&&&&\\
		\cmidrule{2-8}
		&Universal Transf. \citep{dehghani2019universal}&A&\ape&\mr{1}{\no}&\mr{1}{\yes}&\mr{1}{\yes}&\mr{1}{$0$} \\
		\cmidrule{2-8}
		&Di-SAN \citep{shen2018disan}&\mr{3}{R}&\mr{3}{\mam}&\mr{3}{\no}&\mr{3}{\yes}&\mr{3}{\yes}&$0$\\
		&TENER \citep{yan2019tener}&&&&&& $2dlh$ \\
		&Rotary \citep{su2021roformer}&&&&&&$0$\\
		\cmidrule{2-8}
		\multirow{3}{0.5cm}{\rotatebox{90}{\textbf{Trees}}}
		&SPR-abs \citep{wang2019self}&A&\ape&\no&\no&\yes&$0$\\
		&SPR-rel \citep{wang2019self}&R&\mam&\yes&\no&\no&$2(2\tmax{}+1)d$\\
		&TPE \citep{shiv2019novel}&A&\ape&\yes&\no&\no&$\frac{d}{\dmax{}}$\\  
		\cmidrule{2-8}
		\multirow{4}{0.5cm}{\rotatebox{90}{\textbf{Graphs}}}
		&Struct.\ Transformer \citep{zhu-etal-2019-modeling}&\mr{2}{R}&\mr{2}{\mam}&\mr{2}{\yes}&\mr{2}{\yes}&\mr{2}{\yes}&$5d^2+(d+1)d_r$\\
		&Graph Transformer \citep{cai-lam-2020-graph}&&&&&&$7d^2+3d$\\
		\cmidrule{2-8}
		&Graformer \citep{schmitt2021modeling}&R&\mam&\yes&\yes&\no&$2(\dmax + 1)h$\\
		\cmidrule{2-8}
		&Graph Transformer \citep{dwivedi2021generalization}&A&\mr{2}{\ape}&\mr{2}{\no}&\mr{2}{\no}&\mr{2}{\yes}&\mr{2}{$0$}\\
		&\textsc{graph-bert} \citep{zhang-etal-2020-graph}&B&&&&&\\
		\bottomrule
	\end{tabular}
	\caption{Comparison according to several criteria: \posType{} = \Disttype{} (\textbf{A}bsolute, \textbf{R}elative, or \textbf{B}oth); \inj{} = Injection method (\ape{} or \mam{}); \lea{} = Are the position representations learned during training?; \rec{} = Is position information recurring at each layer vs.\ only before first layer?; \eat{\sha{} = Are the same position embeddings reused in each layer? (only makes sense together with \rec{}); }\unb{} = Can the position model generalize to longer inputs than a fixed value?; \param{} = Number of parameters introduced by the position model (with $d$ hidden dimension, $h$ number of attention heads, $\tmax{}$ longest considered sequence length, $l$ number of layers, $\dmax{}$ biggest length of all shortest paths in a graph). Approaches are clustered to avoid repetition and otherwise listed in the same order as discussed in the text. The \other{} symbol means that an entry does not fit into our categories. Note that a model as a whole can combine different position models while this comparison focuses on the respective novel part(s).}
	\tablabel{comparison}
\end{table}

%% file: posover.bbl
\begin{thebibliography}{58}
\providecommand{\natexlab}[1]{#1}
\providecommand{\url}[1]{\texttt{#1}}
\expandafter\ifx\csname urlstyle\endcsname\relax
  \providecommand{\doi}[1]{doi: #1}\else
  \providecommand{\doi}{doi: \begingroup \urlstyle{rm}\Url}\fi

\bibitem[Artetxe et~al.(2020)Artetxe, Ruder, and
  Yogatama]{artetxe2020crosslingual}
Mikel Artetxe, Sebastian Ruder, and Dani Yogatama.
\newblock On the cross-lingual transferability of monolingual representations.
\newblock In Dan Jurafsky, Joyce Chai, Natalie Schluter, and Joel~R. Tetreault
  (eds.), \emph{Proceedings of the 58th Annual Meeting of the Association for
  Computational Linguistics, {ACL} 2020, Online, July 5-10, 2020}, pp.\
  4623--4637. Association for Computational Linguistics, 2020.
\newblock URL \url{https://www.aclweb.org/anthology/2020.acl-main.421/}.

\bibitem[Ba et~al.(2016)Ba, Kiros, and Hinton]{lei2016layer}
Lei~Jimmy Ba, Jamie~Ryan Kiros, and Geoffrey~E. Hinton.
\newblock Layer normalization.
\newblock \emph{Computing Research Repository}, abs/1607.06450, 2016.
\newblock URL \url{http://arxiv.org/abs/1607.06450}.

\bibitem[Bahdanau et~al.(2015)Bahdanau, Cho, and Bengio]{bahdanau2015neural}
Dzmitry Bahdanau, Kyunghyun Cho, and Yoshua Bengio.
\newblock Neural machine translation by jointly learning to align and
  translate.
\newblock In Yoshua Bengio and Yann LeCun (eds.), \emph{3rd International
  Conference on Learning Representations, {ICLR} 2015, San Diego, CA, USA, May
  7-9, 2015, Conference Track Proceedings}, 2015.
\newblock URL \url{http://arxiv.org/abs/1409.0473}.

\bibitem[Bao et~al.(2019)Bao, Zhou, Feng, Wang, Huang, Chen, and
  Li]{bao2019non}
Yu~Bao, Hao Zhou, Jiangtao Feng, Mingxuan Wang, Shujian Huang, Jiajun Chen, and
  Lei Li.
\newblock Non-autoregressive transformer by position learning.
\newblock \emph{Computing Research Repository}, abs/1911.10677, 2019.
\newblock URL \url{http://arxiv.org/abs/1911.10677}.

\bibitem[Brown et~al.(2020)Brown, Mann, Ryder, Subbiah, Kaplan, Dhariwal,
  Neelakantan, Shyam, Sastry, Askell, Agarwal, Herbert{-}Voss, Krueger,
  Henighan, Child, Ramesh, Ziegler, Wu, Winter, Hesse, Chen, Sigler, Litwin,
  Gray, Chess, Clark, Berner, McCandlish, Radford, Sutskever, and
  Amodei]{brown2020language}
Tom~B. Brown, Benjamin Mann, Nick Ryder, Melanie Subbiah, Jared Kaplan,
  Prafulla Dhariwal, Arvind Neelakantan, Pranav Shyam, Girish Sastry, Amanda
  Askell, Sandhini Agarwal, Ariel Herbert{-}Voss, Gretchen Krueger, Tom
  Henighan, Rewon Child, Aditya Ramesh, Daniel~M. Ziegler, Jeffrey Wu, Clemens
  Winter, Christopher Hesse, Mark Chen, Eric Sigler, Mateusz Litwin, Scott
  Gray, Benjamin Chess, Jack Clark, Christopher Berner, Sam McCandlish, Alec
  Radford, Ilya Sutskever, and Dario Amodei.
\newblock Language models are few-shot learners.
\newblock In Hugo Larochelle, Marc'Aurelio Ranzato, Raia Hadsell,
  Maria{-}Florina Balcan, and Hsuan{-}Tien Lin (eds.), \emph{Advances in Neural
  Information Processing Systems 33: Annual Conference on Neural Information
  Processing Systems 2020, NeurIPS 2020, December 6-12, 2020, virtual}, 2020.
\newblock URL
  \url{https://proceedings.neurips.cc/paper/2020/hash/1457c0d6bfcb4967418bfb8ac142f64a-Abstract.html}.

\bibitem[Cai \& Lam(2020)Cai and Lam]{cai-lam-2020-graph}
Deng Cai and Wai Lam.
\newblock Graph transformer for graph-to-sequence learning.
\newblock \emph{AAAI Conference on Artificial Intelligence}, 2020.
\newblock URL \url{https://aaai.org/Papers/AAAI/2020GB/AAAI-CaiD.6741.pdf}.

\bibitem[Chang et~al.(2021)Chang, Xu, Xu, and Tu]{chang-etal-2021-convolutions}
Tyler Chang, Yifan Xu, Weijian Xu, and Zhuowen Tu.
\newblock Convolutions and self-attention: {R}e-interpreting relative positions
  in pre-trained language models.
\newblock In \emph{Proceedings of the 59th Annual Meeting of the Association
  for Computational Linguistics and the 11th International Joint Conference on
  Natural Language Processing (Volume 1: Long Papers)}, pp.\  4322--4333,
  Online, August 2021. Association for Computational Linguistics.
\newblock \doi{10.18653/v1/2021.acl-long.333}.
\newblock URL \url{https://aclanthology.org/2021.acl-long.333}.

\bibitem[Chen et~al.(2021)Chen, Tsai, Bhojanapalli, Chung, Chang, and
  Ferng]{chen2021demystifying}
Pu{-}Chin Chen, Henry Tsai, Srinadh Bhojanapalli, Hyung~Won Chung, Yin{-}Wen
  Chang, and Chun{-}Sung Ferng.
\newblock Demystifying the better performance of position encoding variants for
  transformer.
\newblock \emph{Computing Research Repository}, abs/2104.08698, 2021.
\newblock URL \url{https://arxiv.org/abs/2104.08698}.

\bibitem[Cho et~al.(2014)Cho, van Merrienboer, G{\"{u}}l{\c{c}}ehre, Bahdanau,
  Bougares, Schwenk, and Bengio]{cho2014learning}
Kyunghyun Cho, Bart van Merrienboer, {\c{C}}aglar G{\"{u}}l{\c{c}}ehre, Dzmitry
  Bahdanau, Fethi Bougares, Holger Schwenk, and Yoshua Bengio.
\newblock Learning phrase representations using {RNN} encoder-decoder for
  statistical machine translation.
\newblock In Alessandro Moschitti, Bo~Pang, and Walter Daelemans (eds.),
  \emph{Proceedings of the 2014 Conference on Empirical Methods in Natural
  Language Processing, {EMNLP} 2014, October 25-29, 2014, Doha, Qatar, {A}
  meeting of SIGDAT, a Special Interest Group of the {ACL}}, pp.\  1724--1734.
  {ACL}, 2014.
\newblock \doi{10.3115/v1/d14-1179}.
\newblock URL \url{https://doi.org/10.3115/v1/d14-1179}.

\bibitem[Choromanski et~al.(2021)Choromanski, Likhosherstov, Dohan, Song, Gane,
  Sarlos, Hawkins, Davis, Mohiuddin, Kaiser, Belanger, Colwell, and
  Weller]{choromanski-etal-2021-rethinking}
Krzysztof~Marcin Choromanski, Valerii Likhosherstov, David Dohan, Xingyou Song,
  Andreea Gane, Tamas Sarlos, Peter Hawkins, Jared~Quincy Davis, Afroz
  Mohiuddin, Lukasz Kaiser, David~Benjamin Belanger, Lucy~J Colwell, and Adrian
  Weller.
\newblock Rethinking attention with performers.
\newblock In \emph{International Conference on Learning Representations}, 2021.
\newblock URL \url{https://openreview.net/forum?id=Ua6zuk0WRH}.

\bibitem[Conneau et~al.(2020)Conneau, Khandelwal, Goyal, Chaudhary, Wenzek,
  Guzm{\'{a}}n, Grave, Ott, Zettlemoyer, and Stoyanov]{conneau2020unsupervised}
Alexis Conneau, Kartikay Khandelwal, Naman Goyal, Vishrav Chaudhary, Guillaume
  Wenzek, Francisco Guzm{\'{a}}n, Edouard Grave, Myle Ott, Luke Zettlemoyer,
  and Veselin Stoyanov.
\newblock Unsupervised cross-lingual representation learning at scale.
\newblock In Dan Jurafsky, Joyce Chai, Natalie Schluter, and Joel~R. Tetreault
  (eds.), \emph{Proceedings of the 58th Annual Meeting of the Association for
  Computational Linguistics, {ACL} 2020, Online, July 5-10, 2020}, pp.\
  8440--8451. Association for Computational Linguistics, 2020.
\newblock \doi{10.18653/v1/2020.acl-main.747}.
\newblock URL \url{https://doi.org/10.18653/v1/2020.acl-main.747}.

\bibitem[Dai et~al.(2019)Dai, Yang, Yang, Carbonell, Le, and
  Salakhutdinov]{dai-etal-2019-transformer}
Zihang Dai, Zhilin Yang, Yiming Yang, Jaime Carbonell, Quoc Le, and Ruslan
  Salakhutdinov.
\newblock Transformer-{XL}: Attentive language models beyond a fixed-length
  context.
\newblock In \emph{Proceedings of the 57th Annual Meeting of the Association
  for Computational Linguistics}, pp.\  2978--2988, Florence, Italy, July 2019.
  Association for Computational Linguistics.
\newblock \doi{10.18653/v1/P19-1285}.
\newblock URL \url{https://www.aclweb.org/anthology/P19-1285}.

\bibitem[Dehghani et~al.(2019)Dehghani, Gouws, Vinyals, Uszkoreit, and
  Kaiser]{dehghani2019universal}
Mostafa Dehghani, Stephan Gouws, Oriol Vinyals, Jakob Uszkoreit, and Lukasz
  Kaiser.
\newblock Universal transformers.
\newblock In \emph{7th International Conference on Learning Representations,
  {ICLR} 2019, New Orleans, LA, USA, May 6-9, 2019}. OpenReview.net, 2019.
\newblock URL \url{https://openreview.net/forum?id=HyzdRiR9Y7}.

\bibitem[Devlin et~al.(2019)Devlin, Chang, Lee, and
  Toutanova]{devlin-etal-2019-bert}
Jacob Devlin, Ming-Wei Chang, Kenton Lee, and Kristina Toutanova.
\newblock {BERT}: Pre-training of deep bidirectional transformers for language
  understanding.
\newblock In \emph{Proceedings of the 2019 Conference of the North {A}merican
  Chapter of the Association for Computational Linguistics: Human Language
  Technologies, Volume 1 (Long and Short Papers)}, pp.\  4171--4186,
  Minneapolis, Minnesota, June 2019. Association for Computational Linguistics.
\newblock \doi{10.18653/v1/N19-1423}.
\newblock URL \url{https://www.aclweb.org/anthology/N19-1423}.

\bibitem[Ding et~al.(2020)Ding, Wang, and Tao]{ding2020self}
Liang Ding, Longyue Wang, and Dacheng Tao.
\newblock Self-attention with cross-lingual position representation.
\newblock In Dan Jurafsky, Joyce Chai, Natalie Schluter, and Joel~R. Tetreault
  (eds.), \emph{Proceedings of the 58th Annual Meeting of the Association for
  Computational Linguistics, {ACL} 2020, Online, July 5-10, 2020}, pp.\
  1679--1685. Association for Computational Linguistics, 2020.
\newblock \doi{10.18653/v1/2020.acl-main.153}.
\newblock URL \url{https://doi.org/10.18653/v1/2020.acl-main.153}.

\bibitem[Dufter(2021)]{dufter2021distributed}
Philipp Dufter.
\newblock \emph{Distributed representations for multilingual language
  processing}.
\newblock PhD thesis, Ludwig-Maximilians-Universität München, 2021.

\bibitem[Dufter et~al.(2020)Dufter, Schmitt, and
  Sch{\"u}tze]{dufter2020increasing}
Philipp Dufter, Martin Schmitt, and Hinrich Sch{\"u}tze.
\newblock Increasing learning efficiency of self-attention networks through
  direct position interactions, learnable temperature, and convoluted
  attention.
\newblock In \emph{Proceedings of the 28th International Conference on
  Computational Linguistics}, pp.\  3630--3636, Barcelona, Spain (Online),
  December 2020. International Committee on Computational Linguistics.
\newblock \doi{10.18653/v1/2020.coling-main.324}.
\newblock URL \url{https://www.aclweb.org/anthology/2020.coling-main.324}.

\bibitem[Dwivedi \& Bresson(2021)Dwivedi and
  Bresson]{dwivedi2021generalization}
Vijay~Prakash Dwivedi and Xavier Bresson.
\newblock A generalization of transformer networks to graphs.
\newblock \emph{AAAI Workshop on Deep Learning on Graphs: Methods and
  Applications}, 2021.

\bibitem[Gehring et~al.(2017)Gehring, Auli, Grangier, Yarats, and
  Dauphin]{gehring2017convolutional}
Jonas Gehring, Michael Auli, David Grangier, Denis Yarats, and Yann~N. Dauphin.
\newblock Convolutional sequence to sequence learning.
\newblock In Doina Precup and Yee~Whye Teh (eds.), \emph{Proceedings of the
  34th International Conference on Machine Learning, {ICML} 2017, Sydney, NSW,
  Australia, 6-11 August 2017}, volume~70 of \emph{Proceedings of Machine
  Learning Research}, pp.\  1243--1252. {PMLR}, 2017.
\newblock URL \url{http://proceedings.mlr.press/v70/gehring17a.html}.

\bibitem[Harris et~al.(2020)Harris, Millman, van~der Walt, Gommers, Virtanen,
  Cournapeau, Wieser, Taylor, Berg, Smith, Kern, Picus, Hoyer, van Kerkwijk,
  Brett, Haldane, del R{\'{i}}o, Wiebe, Peterson, G{\'{e}}rard-Marchant,
  Sheppard, Reddy, Weckesser, Abbasi, Gohlke, and Oliphant]{harris2020array}
Charles~R. Harris, K.~Jarrod Millman, St{\'{e}}fan~J. van~der Walt, Ralf
  Gommers, Pauli Virtanen, David Cournapeau, Eric Wieser, Julian Taylor,
  Sebastian Berg, Nathaniel~J. Smith, Robert Kern, Matti Picus, Stephan Hoyer,
  Marten~H. van Kerkwijk, Matthew Brett, Allan Haldane, Jaime~Fern{\'{a}}ndez
  del R{\'{i}}o, Mark Wiebe, Pearu Peterson, Pierre G{\'{e}}rard-Marchant,
  Kevin Sheppard, Tyler Reddy, Warren Weckesser, Hameer Abbasi, Christoph
  Gohlke, and Travis~E. Oliphant.
\newblock Array programming with {NumPy}.
\newblock \emph{Nature}, 585\penalty0 (7825):\penalty0 357--362, September
  2020.
\newblock \doi{10.1038/s41586-020-2649-2}.
\newblock URL \url{https://doi.org/10.1038/s41586-020-2649-2}.

\bibitem[He et~al.(2021)He, Liu, Gao, and Chen]{he2020deberta}
Pengcheng He, Xiaodong Liu, Jianfeng Gao, and Weizhu Chen.
\newblock De{BERT}a: decoding-enhanced {BERT} with disentangled attention.
\newblock In \emph{9th International Conference on Learning Representations,
  {ICLR} 2021, Virtual Event, Austria, May 3-7, 2021}. OpenReview.net, 2021.
\newblock URL \url{https://openreview.net/forum?id=XPZIaotutsD}.

\bibitem[Howard \& Ruder(2018)Howard and Ruder]{howard-ruder-2018-universal}
Jeremy Howard and Sebastian Ruder.
\newblock Universal language model fine-tuning for text classification.
\newblock In \emph{Proceedings of the 56th Annual Meeting of the Association
  for Computational Linguistics (Volume 1: Long Papers)}, pp.\  328--339,
  Melbourne, Australia, July 2018. Association for Computational Linguistics.
\newblock \doi{10.18653/v1/P18-1031}.
\newblock URL \url{https://www.aclweb.org/anthology/P18-1031}.

\bibitem[Huang et~al.(2020)Huang, Liang, Xu, and Xiang]{huang2020improve}
Zhiheng Huang, Davis Liang, Peng Xu, and Bing Xiang.
\newblock Improve transformer models with better relative position embeddings.
\newblock In Trevor Cohn, Yulan He, and Yang Liu (eds.), \emph{Proceedings of
  the 2020 Conference on Empirical Methods in Natural Language Processing:
  Findings, {EMNLP} 2020, Online Event, 16-20 November 2020}, pp.\  3327--3335.
  Association for Computational Linguistics, 2020.
\newblock \doi{10.18653/v1/2020.findings-emnlp.298}.
\newblock URL \url{https://doi.org/10.18653/v1/2020.findings-emnlp.298}.

\bibitem[Kalchbrenner et~al.(2014)Kalchbrenner, Grefenstette, and
  Blunsom]{kalchbrenner-etal-2014-convolutional}
Nal Kalchbrenner, Edward Grefenstette, and Phil Blunsom.
\newblock A convolutional neural network for modelling sentences.
\newblock In \emph{Proceedings of the 52nd Annual Meeting of the Association
  for Computational Linguistics (Volume 1: Long Papers)}, pp.\  655--665,
  Baltimore, Maryland, June 2014. Association for Computational Linguistics.
\newblock \doi{10.3115/v1/P14-1062}.
\newblock URL \url{https://www.aclweb.org/anthology/P14-1062}.

\bibitem[Ke et~al.(2021)Ke, He, and Liu]{ke2020rethinking}
Guolin Ke, Di~He, and Tie{-}Yan Liu.
\newblock Rethinking positional encoding in language pre-training.
\newblock In \emph{9th International Conference on Learning Representations,
  {ICLR} 2021, Virtual Event, Austria, May 3-7, 2021}. OpenReview.net, 2021.
\newblock URL \url{https://openreview.net/forum?id=09-528y2Fgf}.

\bibitem[Kitaev et~al.(2020)Kitaev, Kaiser, and Levskaya]{kitaev2020reformer}
Nikita Kitaev, Lukasz Kaiser, and Anselm Levskaya.
\newblock Reformer: The efficient transformer.
\newblock In \emph{8th International Conference on Learning Representations,
  {ICLR} 2020, Addis Ababa, Ethiopia, April 26-30, 2020}. OpenReview.net, 2020.
\newblock URL \url{https://openreview.net/forum?id=rkgNKkHtvB}.

\bibitem[Li et~al.(2019)Li, Wang, Liu, Tang, Lei, and Li]{li2019augmented}
Hailiang Li, Adele Y.~C. Wang, Yang Liu, Du~Tang, Zhibin Lei, and Wenye Li.
\newblock An augmented transformer architecture for natural language generation
  tasks.
\newblock In Panagiotis Papapetrou, Xueqi Cheng, and Qing He (eds.), \emph{2019
  International Conference on Data Mining Workshops, {ICDM} Workshops 2019,
  Beijing, China, November 8-11, 2019}, pp.\  1--7. {IEEE}, 2019.
\newblock \doi{10.1109/ICDMW48858.2019.9024754}.
\newblock URL \url{https://doi.org/10.1109/ICDMW48858.2019.9024754}.

\bibitem[Likhomanenko et~al.(2021)Likhomanenko, Xu, Collobert, Synnaeve, and
  Rogozhnikov]{likhomanenko2021cape}
Tatiana Likhomanenko, Qiantong Xu, Ronan Collobert, Gabriel Synnaeve, and Alex
  Rogozhnikov.
\newblock {CAPE:} encoding relative positions with continuous augmented
  positional embeddings.
\newblock \emph{Computing Research Repository}, abs/2106.03143, 2021.
\newblock URL \url{https://arxiv.org/abs/2106.03143}.

\bibitem[Liu et~al.(2021{\natexlab{a}})Liu, Niehues, Cross, Guzm{\'{a}}n, and
  Li]{liu2020improving}
Danni Liu, Jan Niehues, James Cross, Francisco Guzm{\'{a}}n, and Xian Li.
\newblock Improving zero-shot translation by disentangling positional
  information.
\newblock In Chengqing Zong, Fei Xia, Wenjie Li, and Roberto Navigli (eds.),
  \emph{Proceedings of the 59th Annual Meeting of the Association for
  Computational Linguistics and the 11th International Joint Conference on
  Natural Language Processing, {ACL/IJCNLP} 2021, (Volume 1: Long Papers),
  Virtual Event, August 1-6, 2021}, pp.\  1259--1273. Association for
  Computational Linguistics, 2021{\natexlab{a}}.
\newblock \doi{10.18653/v1/2021.acl-long.101}.
\newblock URL \url{https://doi.org/10.18653/v1/2021.acl-long.101}.

\bibitem[Liu et~al.(2020)Liu, Yu, Dhillon, and Hsieh]{liu2020learning}
Xuanqing Liu, Hsiang{-}Fu Yu, Inderjit~S. Dhillon, and Cho{-}Jui Hsieh.
\newblock Learning to encode position for transformer with continuous dynamical
  model.
\newblock In \emph{Proceedings of the 37th International Conference on Machine
  Learning, {ICML} 2020, 13-18 July 2020, Virtual Event}, volume 119 of
  \emph{Proceedings of Machine Learning Research}, pp.\  6327--6335. {PMLR},
  2020.
\newblock URL \url{http://proceedings.mlr.press/v119/liu20n.html}.

\bibitem[Liu et~al.(2019)Liu, Ott, Goyal, Du, Joshi, Chen, Levy, Lewis,
  Zettlemoyer, and Stoyanov]{liu2019roberta}
Yinhan Liu, Myle Ott, Naman Goyal, Jingfei Du, Mandar Joshi, Danqi Chen, Omer
  Levy, Mike Lewis, Luke Zettlemoyer, and Veselin Stoyanov.
\newblock Roberta: {A} robustly optimized {BERT} pretraining approach.
\newblock \emph{Computing Research Repository}, abs/1907.11692, 2019.
\newblock URL \url{http://arxiv.org/abs/1907.11692}.

\bibitem[Liu et~al.(2021{\natexlab{b}})Liu, Winata, Cahyawijaya, Madotto, Lin,
  and Fung]{liu2020do}
Zihan Liu, Genta~Indra Winata, Samuel Cahyawijaya, Andrea Madotto, Zhaojiang
  Lin, and Pascale Fung.
\newblock On the importance of word order information in cross-lingual sequence
  labeling.
\newblock In \emph{Thirty-Fifth {AAAI} Conference on Artificial Intelligence,
  {AAAI} 2021, Thirty-Third Conference on Innovative Applications of Artificial
  Intelligence, {IAAI} 2021, The Eleventh Symposium on Educational Advances in
  Artificial Intelligence, {EAAI} 2021, Virtual Event, February 2-9, 2021},
  pp.\  13461--13469. {AAAI} Press, 2021{\natexlab{b}}.
\newblock URL \url{https://ojs.aaai.org/index.php/AAAI/article/view/17588}.

\bibitem[Liutkus et~al.(2021)Liutkus, C{\'i}fka, Wu, {\c S}im{\c s}ekli, Yang,
  and Richard]{liutkus-etal-2021-relative}
Antoine Liutkus, Ond{\v r}ej C{\'i}fka, Shih-Lun Wu, Umut {\c S}im{\c s}ekli,
  Yi-Hsuan Yang, and Ga{\"e}l Richard.
\newblock Relative positional encoding for {Transformers} with linear
  complexity.
\newblock In Marina Meila and Tong Zhang (eds.), \emph{Proceedings of the 38th
  International Conference on Machine Learning}, volume 139 of
  \emph{Proceedings of Machine Learning Research}, pp.\  7067--7079. PMLR,
  18--24 Jul 2021.
\newblock URL \url{http://proceedings.mlr.press/v139/liutkus21a.html}.

\bibitem[Neishi \& Yoshinaga(2019)Neishi and Yoshinaga]{neishi2019relation}
Masato Neishi and Naoki Yoshinaga.
\newblock On the relation between position information and sentence length in
  neural machine translation.
\newblock In Mohit Bansal and Aline Villavicencio (eds.), \emph{Proceedings of
  the 23rd Conference on Computational Natural Language Learning, CoNLL 2019,
  Hong Kong, China, November 3-4, 2019}, pp.\  328--338. Association for
  Computational Linguistics, 2019.
\newblock \doi{10.18653/v1/K19-1031}.
\newblock URL \url{https://doi.org/10.18653/v1/K19-1031}.

\bibitem[Oka et~al.(2020)Oka, Chousa, Sudoh, and
  Nakamura]{oka-etal-2020-incorporating}
Yui Oka, Katsuki Chousa, Katsuhito Sudoh, and Satoshi Nakamura.
\newblock Incorporating noisy length constraints into transformer with
  length-aware positional encodings.
\newblock In \emph{Proceedings of the 28th International Conference on
  Computational Linguistics}, pp.\  3580--3585, Barcelona, Spain (Online),
  December 2020. International Committee on Computational Linguistics.
\newblock URL \url{https://www.aclweb.org/anthology/2020.coling-main.319}.

\bibitem[Peters et~al.(2018)Peters, Neumann, Iyyer, Gardner, Clark, Lee, and
  Zettlemoyer]{peters-etal-2018-deep}
Matthew Peters, Mark Neumann, Mohit Iyyer, Matt Gardner, Christopher Clark,
  Kenton Lee, and Luke Zettlemoyer.
\newblock Deep contextualized word representations.
\newblock In \emph{Proceedings of the 2018 Conference of the North {A}merican
  Chapter of the Association for Computational Linguistics: Human Language
  Technologies, Volume 1 (Long Papers)}, pp.\  2227--2237, New Orleans,
  Louisiana, June 2018. Association for Computational Linguistics.
\newblock \doi{10.18653/v1/N18-1202}.
\newblock URL \url{https://www.aclweb.org/anthology/N18-1202}.

\bibitem[Press et~al.(2021)Press, Smith, and Lewis]{press2020shortformer}
Ofir Press, Noah~A. Smith, and Mike Lewis.
\newblock Shortformer: Better language modeling using shorter inputs.
\newblock In Chengqing Zong, Fei Xia, Wenjie Li, and Roberto Navigli (eds.),
  \emph{Proceedings of the 59th Annual Meeting of the Association for
  Computational Linguistics and the 11th International Joint Conference on
  Natural Language Processing, {ACL/IJCNLP} 2021, (Volume 1: Long Papers),
  Virtual Event, August 1-6, 2021}, pp.\  5493--5505. Association for
  Computational Linguistics, 2021.
\newblock \doi{10.18653/v1/2021.acl-long.427}.
\newblock URL \url{https://doi.org/10.18653/v1/2021.acl-long.427}.

\bibitem[Radford et~al.(2019)Radford, Wu, Child, Luan, Amodei, and
  Sutskever]{radford2019language}
Alec Radford, Jeffrey Wu, Rewon Child, David Luan, Dario Amodei, and Ilya
  Sutskever.
\newblock Language models are unsupervised multitask learners.
\newblock \emph{OpenAI blog}, 1\penalty0 (8):\penalty0 9, 2019.

\bibitem[Raffel et~al.(2020)Raffel, Shazeer, Roberts, Lee, Narang, Matena,
  Zhou, Li, and Liu]{raffel2020exploring}
Colin Raffel, Noam Shazeer, Adam Roberts, Katherine Lee, Sharan Narang, Michael
  Matena, Yanqi Zhou, Wei Li, and Peter~J. Liu.
\newblock Exploring the limits of transfer learning with a unified text-to-text
  transformer.
\newblock \emph{Journal of Machine Learning Research}, 21:\penalty0
  140:1--140:67, 2020.
\newblock URL \url{http://jmlr.org/papers/v21/20-074.html}.

\bibitem[Rosendahl et~al.(2019)Rosendahl, Tran, Wang, and
  Ney]{rosendahl2019analysis}
Jan Rosendahl, Viet Anh~Khoa Tran, Weiyue Wang, and Hermann Ney.
\newblock Analysis of positional encodings for neural machine translation.
\newblock \emph{IWSLT, Hong Kong, China}, 2019.
\newblock URL
  \url{https://www-i6.informatik.rwth-aachen.de/publications/download/1132/RosendahlJanTranVietAnhKhoaWangWeiyueNeyHermann--AnalysisofPositionalEncodingsforNeuralMachineTranslation--2019.pdf}.

\bibitem[Schmitt et~al.(2021)Schmitt, Ribeiro, Dufter, Gurevych, and
  Sch{\"u}tze]{schmitt2021modeling}
Martin Schmitt, Leonardo F.~R. Ribeiro, Philipp Dufter, Iryna Gurevych, and
  Hinrich Sch{\"u}tze.
\newblock Modeling graph structure via relative position for text generation
  from knowledge graphs.
\newblock In \emph{Proceedings of the Fifteenth Workshop on Graph-Based Methods
  for Natural Language Processing (TextGraphs-15)}, pp.\  10--21, Mexico City,
  Mexico, June 2021. Association for Computational Linguistics.
\newblock URL \url{https://aclanthology.org/2021.textgraphs-1.2}.

\bibitem[Shaw et~al.(2018)Shaw, Uszkoreit, and Vaswani]{shaw2018self}
Peter Shaw, Jakob Uszkoreit, and Ashish Vaswani.
\newblock Self-attention with relative position representations.
\newblock In Marilyn~A. Walker, Heng Ji, and Amanda Stent (eds.),
  \emph{Proceedings of the 2018 Conference of the North American Chapter of the
  Association for Computational Linguistics: Human Language Technologies,
  NAACL-HLT, New Orleans, Louisiana, USA, June 1-6, 2018, Volume 2 (Short
  Papers)}, pp.\  464--468. Association for Computational Linguistics, 2018.
\newblock \doi{10.18653/v1/n18-2074}.
\newblock URL \url{https://doi.org/10.18653/v1/n18-2074}.

\bibitem[Shen et~al.(2018)Shen, Zhou, Long, Jiang, Pan, and
  Zhang]{shen2018disan}
Tao Shen, Tianyi Zhou, Guodong Long, Jing Jiang, Shirui Pan, and Chengqi Zhang.
\newblock Disan: Directional self-attention network for rnn/cnn-free language
  understanding.
\newblock In Sheila~A. McIlraith and Kilian~Q. Weinberger (eds.),
  \emph{Proceedings of the Thirty-Second {AAAI} Conference on Artificial
  Intelligence, (AAAI-18), the 30th innovative Applications of Artificial
  Intelligence (IAAI-18), and the 8th {AAAI} Symposium on Educational Advances
  in Artificial Intelligence (EAAI-18), New Orleans, Louisiana, USA, February
  2-7, 2018}, pp.\  5446--5455. {AAAI} Press, 2018.
\newblock URL
  \url{https://www.aaai.org/ocs/index.php/AAAI/AAAI18/paper/view/16126}.

\bibitem[Shiv \& Quirk(2019)Shiv and Quirk]{shiv2019novel}
Vighnesh~Leonardo Shiv and Chris Quirk.
\newblock Novel positional encodings to enable tree-based transformers.
\newblock In Hanna~M. Wallach, Hugo Larochelle, Alina Beygelzimer, Florence
  d'Alch{\'{e}}{-}Buc, Emily~B. Fox, and Roman Garnett (eds.), \emph{Advances
  in Neural Information Processing Systems 32: Annual Conference on Neural
  Information Processing Systems 2019, NeurIPS 2019, 8-14 December 2019,
  Vancouver, BC, Canada}, pp.\  12058--12068, 2019.
\newblock URL
  \url{http://papers.nips.cc/paper/9376-novel-positional-encodings-to-enable-tree-based-transformers}.

\bibitem[Su et~al.(2021)Su, Lu, Pan, Wen, and Liu]{su2021roformer}
Jianlin Su, Yu~Lu, Shengfeng Pan, Bo~Wen, and Yunfeng Liu.
\newblock Roformer: Enhanced transformer with rotary position embedding.
\newblock \emph{Computing Research Repository}, abs/2104.09864, 2021.
\newblock URL \url{https://arxiv.org/abs/2104.09864}.

\bibitem[Takase \& Okazaki(2019)Takase and
  Okazaki]{takase-okazaki-2019-positional}
Sho Takase and Naoaki Okazaki.
\newblock Positional encoding to control output sequence length.
\newblock In \emph{Proceedings of the 2019 Conference of the North {A}merican
  Chapter of the Association for Computational Linguistics: Human Language
  Technologies, Volume 1 (Long and Short Papers)}, pp.\  3999--4004,
  Minneapolis, Minnesota, June 2019. Association for Computational Linguistics.
\newblock \doi{10.18653/v1/N19-1401}.
\newblock URL \url{https://www.aclweb.org/anthology/N19-1401}.

\bibitem[Tay et~al.(2021)Tay, Dehghani, Abnar, Shen, Bahri, Pham, Rao, Yang,
  Ruder, and Metzler]{tay-etal-2021-long}
Yi~Tay, Mostafa Dehghani, Samira Abnar, Yikang Shen, Dara Bahri, Philip Pham,
  Jinfeng Rao, Liu Yang, Sebastian Ruder, and Donald Metzler.
\newblock Long range arena : A benchmark for efficient transformers.
\newblock In \emph{International Conference on Learning Representations}, 2021.
\newblock URL \url{https://openreview.net/forum?id=qVyeW-grC2k}.

\bibitem[Vaswani et~al.(2017)Vaswani, Shazeer, Parmar, Uszkoreit, Jones, Gomez,
  Kaiser, and Polosukhin]{vaswani2017attention}
Ashish Vaswani, Noam Shazeer, Niki Parmar, Jakob Uszkoreit, Llion Jones,
  Aidan~N Gomez, \L~ukasz Kaiser, and Illia Polosukhin.
\newblock Attention is all you need.
\newblock In I.~Guyon, U.~V. Luxburg, S.~Bengio, H.~Wallach, R.~Fergus,
  S.~Vishwanathan, and R.~Garnett (eds.), \emph{Advances in Neural Information
  Processing Systems 30}, pp.\  5998--6008. Curran Associates, Inc., 2017.
\newblock URL
  \url{http://papers.nips.cc/paper/7181-attention-is-all-you-need.pdf}.

\bibitem[Wang et~al.(2018)Wang, Singh, Michael, Hill, Levy, and
  Bowman]{wang-etal-2018-glue}
Alex Wang, Amanpreet Singh, Julian Michael, Felix Hill, Omer Levy, and Samuel
  Bowman.
\newblock {GLUE}: A multi-task benchmark and analysis platform for natural
  language understanding.
\newblock In \emph{Proceedings of the 2018 {EMNLP} Workshop {B}lackbox{NLP}:
  Analyzing and Interpreting Neural Networks for {NLP}}, pp.\  353--355,
  Brussels, Belgium, November 2018. Association for Computational Linguistics.
\newblock \doi{10.18653/v1/W18-5446}.
\newblock URL \url{https://www.aclweb.org/anthology/W18-5446}.

\bibitem[Wang et~al.(2020)Wang, Zhao, Lioma, Li, Zhang, and
  Simonsen]{wang2020encoding}
Benyou Wang, Donghao Zhao, Christina Lioma, Qiuchi Li, Peng Zhang, and
  Jakob~Grue Simonsen.
\newblock Encoding word order in complex embeddings.
\newblock In \emph{8th International Conference on Learning Representations,
  {ICLR} 2020, Addis Ababa, Ethiopia, April 26-30, 2020}. OpenReview.net, 2020.
\newblock URL \url{https://openreview.net/forum?id=Hke-WTVtwr}.

\bibitem[Wang et~al.(2021)Wang, Shang, Lioma, Jiang, Yang, Liu, and
  Simonsen]{wang2021on}
Benyou Wang, Lifeng Shang, Christina Lioma, Xin Jiang, Hao Yang, Qun Liu, and
  Jakob~Grue Simonsen.
\newblock On position embeddings in {BERT}.
\newblock In \emph{International Conference on Learning Representations}, 2021.
\newblock URL \url{https://openreview.net/forum?id=onxoVA9FxMw}.

\bibitem[Wang et~al.(2019)Wang, Tu, Wang, and Shi]{wang2019self}
Xing Wang, Zhaopeng Tu, Longyue Wang, and Shuming Shi.
\newblock Self-attention with structural position representations.
\newblock In Kentaro Inui, Jing Jiang, Vincent Ng, and Xiaojun Wan (eds.),
  \emph{Proceedings of the 2019 Conference on Empirical Methods in Natural
  Language Processing and the 9th International Joint Conference on Natural
  Language Processing, {EMNLP-IJCNLP} 2019, Hong Kong, China, November 3-7,
  2019}, pp.\  1403--1409. Association for Computational Linguistics, 2019.
\newblock \doi{10.18653/v1/D19-1145}.
\newblock URL \url{https://doi.org/10.18653/v1/D19-1145}.

\bibitem[Wang \& Chen(2020)Wang and Chen]{wang2020what}
Yu{-}An Wang and Yun{-}Nung Chen.
\newblock What do position embeddings learn? an empirical study of pre-trained
  language model positional encoding.
\newblock In Bonnie Webber, Trevor Cohn, Yulan He, and Yang Liu (eds.),
  \emph{Proceedings of the 2020 Conference on Empirical Methods in Natural
  Language Processing, {EMNLP} 2020, Online, November 16-20, 2020}, pp.\
  6840--6849. Association for Computational Linguistics, 2020.
\newblock \doi{10.18653/v1/2020.emnlp-main.555}.
\newblock URL \url{https://doi.org/10.18653/v1/2020.emnlp-main.555}.

\bibitem[Wu et~al.(2021)Wu, Wu, and Huang]{wu2020datransformer}
Chuhan Wu, Fangzhao Wu, and Yongfeng Huang.
\newblock Da-transformer: Distance-aware transformer.
\newblock In Kristina Toutanova, Anna Rumshisky, Luke Zettlemoyer, Dilek
  Hakkani{-}T{\"{u}}r, Iz~Beltagy, Steven Bethard, Ryan Cotterell, Tanmoy
  Chakraborty, and Yichao Zhou (eds.), \emph{Proceedings of the 2021 Conference
  of the North American Chapter of the Association for Computational
  Linguistics: Human Language Technologies, {NAACL-HLT} 2021, Online, June
  6-11, 2021}, pp.\  2059--2068. Association for Computational Linguistics,
  2021.
\newblock \doi{10.18653/v1/2021.naacl-main.166}.
\newblock URL \url{https://doi.org/10.18653/v1/2021.naacl-main.166}.

\bibitem[Yan et~al.(2019)Yan, Deng, Li, and Qiu]{yan2019tener}
Hang Yan, Bocao Deng, Xiaonan Li, and Xipeng Qiu.
\newblock {TENER:} adapting transformer encoder for named entity recognition.
\newblock \emph{Computing Research Repository}, abs/1911.04474, 2019.
\newblock URL \url{http://arxiv.org/abs/1911.04474}.

\bibitem[Yang et~al.(2019)Yang, Wang, Wong, Chao, and Tu]{yang2019assessing}
Baosong Yang, Longyue Wang, Derek~F. Wong, Lidia~S. Chao, and Zhaopeng Tu.
\newblock Assessing the ability of self-attention networks to learn word order.
\newblock In Anna Korhonen, David~R. Traum, and Llu{\'{\i}}s M{\`{a}}rquez
  (eds.), \emph{Proceedings of the 57th Conference of the Association for
  Computational Linguistics, {ACL} 2019, Florence, Italy, July 28- August 2,
  2019, Volume 1: Long Papers}, pp.\  3635--3644. Association for Computational
  Linguistics, 2019.
\newblock \doi{10.18653/v1/p19-1354}.
\newblock URL \url{https://doi.org/10.18653/v1/p19-1354}.

\bibitem[Zhang et~al.(2020)Zhang, Zhang, Xia, and Sun]{zhang-etal-2020-graph}
Jiawei Zhang, Haopeng Zhang, Congying Xia, and Li~Sun.
\newblock Graph-bert: Only attention is needed for learning graph
  representations.
\newblock \emph{Computing Research Repository}, abs/2001.05140, 2020.
\newblock URL \url{https://arxiv.org/abs/2001.05140}.

\bibitem[Zhu et~al.(2019)Zhu, Li, Zhu, Qian, Zhang, and
  Zhou]{zhu-etal-2019-modeling}
Jie Zhu, Junhui Li, Muhua Zhu, Longhua Qian, Min Zhang, and Guodong Zhou.
\newblock Modeling graph structure in transformer for better {AMR}-to-text
  generation.
\newblock In \emph{Proceedings of the 2019 Conference on Empirical Methods in
  Natural Language Processing and the 9th International Joint Conference on
  Natural Language Processing (EMNLP-IJCNLP)}, pp.\  5459--5468, Hong Kong,
  China, November 2019. Association for Computational Linguistics.
\newblock \doi{10.18653/v1/D19-1548}.
\newblock URL \url{https://www.aclweb.org/anthology/D19-1548}.

\end{thebibliography}
